\documentclass[final]{cvpr}

\usepackage{times}
\usepackage{epsfig}
\usepackage{graphicx}
\usepackage{amsmath}
\usepackage{amssymb}

\usepackage[numbers,sort]{natbib}
\usepackage{graphicx}
\usepackage{mathtools}
\usepackage{tabularx}
\usepackage{multirow}
\usepackage{enumitem}
\usepackage{bbm}
\usepackage{xcolor, colortbl}
\usepackage{wrapfig}
\definecolor{grey}{rgb}{0.9, 0.9, 0.9}
\newcommand{\ccol}{\cellcolor{grey}}

\definecolor{grn}{rgb}{0.1, 0.6, 0.1}
\definecolor{mgt}{rgb}{0.6, 0.1, 0.6}

\newcommand{\sy}[1]{{\color{purple}{#1}}}
\newcommand{\sytodo}[1]{{\color{blue}{#1}}}

\usepackage[pagebackref=true,breaklinks=true,colorlinks,bookmarks=false]{hyperref}

\def\cvprPaperID{2428} % *** Enter the CVPR Paper ID here
\def\confYear{CVPR 2021}
\begin{document}

%%%%%%%%% TITLE
\title{Embedding Transfer with Label Relaxation for Improved Metric Learning}

\author{
Sungyeon Kim$^1$ \qquad
Dongwon Kim$^1$ \qquad
Minsu Cho$^{1,2}$ \qquad
Suha Kwak$^{1,2}$ \\
Dept. of CSE, POSTECH$^1$ \ \ \ \ \ \ \ \ \ \ \ \ \ \ Graduate School of AI, POSTECH$^2$ \\
% Department of CSE$^1$, Graduate School of AI$^2$, POSTECH\\
% POSTECH, Pohang, Korea\\
{\tt\small \{sungyeon.kim, kdwon, mscho, suha.kwak\}@postech.ac.kr}
}

\maketitle

%%% ABSTRACT %%%%%%%%%%%%%%%%%%%%%%%%%%%%%%%%%%%%%%%%%%%%
% !TEX root = cvpr.tex

\begin{abstract}
\label{sec:abstract}
This paper presents a novel method for embedding transfer, a task of transferring knowledge of a learned embedding model to another.
Our method exploits pairwise similarities between samples in the source embedding space as the knowledge, and transfers them through a loss used for learning target embedding models.
To this end, we design a new loss called relaxed contrastive loss, which employs the pairwise similarities as relaxed labels for inter-sample relations.
Our loss provides a rich supervisory signal beyond class equivalence, enables more important pairs to contribute more to training, and imposes no restriction on manifolds of target embedding spaces.
Experiments on metric learning benchmarks demonstrate that our method largely improves performance, or reduces sizes and output dimensions of target models effectively. 
We further show that it can be also used to enhance quality of self-supervised representation and performance of classification models. 
In all the experiments, our method clearly outperforms existing embedding transfer techniques.

\end{abstract}

%%% INTRODUCTION %%%%%%%%%%%%%%%%%%%%%%%%%%%%%%%%%%%%%%%%
% !TEX root = cvpr.tex

\section{Introduction}
\label{sec:intro}

Deep metric learning aims to learn an embedding space where samples of the same class are grouped tightly together. 
Such an embedding space has played important roles in many tasks including image retrieval~\cite{movshovitz2017no,songCVPR16,Sohn_nips2016,kim2019deep,kim2020proxy}, few-shot learning~\cite{snell2017prototypical, sung2018learning, Qiao_2019_ICCV}, zero-shot learning~\cite{Bucher_ECCV_2016, zhang2016zero}, and self-supervised representation learning~\cite{tian2019contrastive, chen2020simple,he2020momentum}.
% \suha{Metric learning, \ie, learning an embedding space where semantically similar samples are grouped together,} has played important roles in many tasks including data retrieval~\cite{movshovitz2017no,songCVPR16,Sohn_nips2016,kim2019deep,kim2020proxy}, few-shot learning~\cite{snell2017prototypical, sung2018learning, Qiao_2019_ICCV}, zero-shot learning~\cite{Bucher_ECCV_2016, zhang2016zero}, and self-supervised representation learning~\cite{tian2019contrastive, chen2020simple,he2020momentum}.
In these tasks, the performance and efficiency of models rely heavily on the quality and dimension of their learned embedding spaces.
To obtain high-quality and compact embedding spaces, previous methods have proposed 
new metric learning losses~\cite{songCVPR16,Sohn_nips2016,Yu_2019_ICCV,wang2019multi,movshovitz2017no,kim2020proxy},
advanced sampling strategies~\cite{sampling_matters,Harwood_2017_ICCV,wang2020cross, ko2020embedding}, 
regularization techniques~\cite{JACOB_2019_ICCV, mohan2020moving}, 
or ensemble models~\cite{Opitz_ICCV_2017,opitz2018deep, ensemble_embedding}.

For the same purpose, we study transferring knowledge of a learned embedding model (source) to another (target), which we call \emph{embedding transfer}.
This task can be considered as a variant of \emph{knowledge distillation}~\cite{hinton2015distilling} that focuses on metric learning instead of classification.
The knowledge captured by the source embedding model can provide rich information beyond class labels such as intra-class variations and degrees of semantic affinity between samples. 
Given a proper way to transfer the knowledge, embedding transfer enables us to improve the performance of target embedding models or compress them, as knowledge distillation does for classification models~\cite{hinton2015distilling,romero2014fitnets,zagoruyko2016paying,yim2017gift,furlanello2018born}. 
% The two main factors of embedding transfer, the type of knowledge and the way to transfer, thus have to be carefully designed for the success of the task.
% Thus, the type of knowledge and the way to transfer have to be carefully designed for the success of embedding transfer.

% In spite of its short history, a few interesting methods have been proposed for embedding transfer.
Existing methods for embedding transfer extract knowledge from a source embedding space in forms of probability distributions of samples~\cite{passalis2018learning}, their geometric relations~\cite{park2019relational,yu2019learning}, or the rank of their similarities~\cite{chen2017darkrank}.
The knowledge is then transferred by forcing target models to approximate those extracted patterns directly in their embedding spaces.
Although these methods shed light on the effective yet less explored approach to enhancing the performance of metric learning, there is still large room for further improvement.
In particular, they fail to utilize detailed inter-sample relations in the source embedding space~\cite{passalis2018learning,chen2017darkrank} 
or blindly accept the transferred knowledge without considering the importance of samples~\cite{park2019relational,yu2019learning}.
% , both of which lead to suboptimal solutions.
% since they often fail to utilize detailed inter-sample relations in the source embedding space~\cite{passalis2018learning,chen2017darkrank},
% and equally handle the samples delivering knowledge~\cite{park2019relational,yu2019learning}, although the sampls are not equally important in practice.

% % ======================================================================
% FIGURE START
\begin{figure*} [!t]
\centering
\includegraphics[width = 0.96 \textwidth]{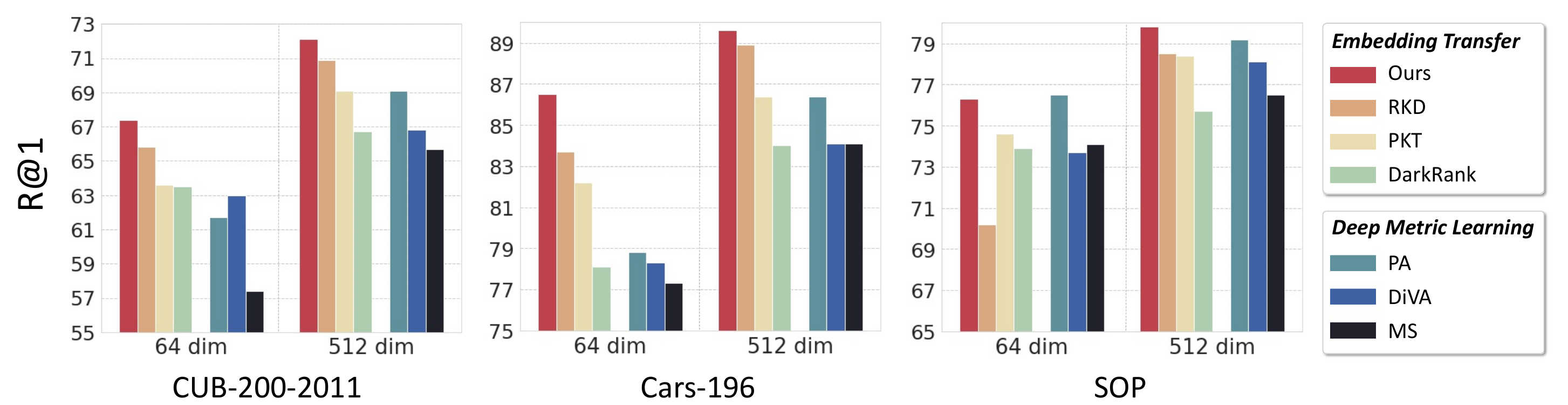}
\caption{
% Accuracy in Recall@1 on CUB-200-2011, Cars-196, and SOP datasets. Our loss enable to achieve the highest accuracy compared to other embedding transfer methods and state-of-the-art of deep metric learning. Detailed results are in Table ~\ref{tab:eval_cub_cars_sop}.
Accuracy in Recall@1 on the three standard benchmarks for deep metric learning. 
All embedding transfer methods adopt Proxy-Anchor (PA)~\cite{kim2020proxy} with 512 dimension as the source model.
% Our method substantially outperforms state-of-the-art metric learning models and other embedding transfer methods. 
% Our method best improves performance of the source model and achieves state of the art when embedding dimension is 512,
Our method achieves the state of the art when embedding dimension is 512, 
and is as competitive as recent metric learning models even with a substantially smaller embedding dimension.
In all experiments, it is superior to other embedding transfer techniques. 
More results can be found in Table~\ref{tab:emb_transfer_comp} and~\ref{tab:dml_cub_cars_sop}. 
} 
\label{fig:teaser}
% \vspace{-2mm}
\end{figure*}
% % ======================================================================

This paper presents a new embedding transfer method that overcomes the above limitations. 
% Our method makes use of pairwise similarities between samples in a source embedding space as the knowledge to be transferred.
Our method defines the knowledge as pairwise similarities between samples in a source embedding space.
Pairwise similarities are useful to characterize an embedding space in detail, thus have been widely used for learning embedding spaces~\cite{Hadsell2006,Schroff2015,Sohn_nips2016,wang2019multi} and identifying underlying manifolds of data~\cite{Cox08_MDS,Tenenbaum00_Isomap}.
Also, they capture detailed inter-sample relations, which are missing in probability distributions~\cite{passalis2018learning} and the rank of similarities~\cite{chen2017darkrank} used as knowledge in previous work.

% For transferring the knowledge, we propose smooth contrastive loss that allows a target embedding space to reflect pairwise semantic similarities observed in the source embedding space.
% Our loss pushes apart or pulls together a pair of samples in a target embedding space as the original contrastive loss does~\cite{Hadsell2006}, but with strength determined by their pairwise similarities in the source embedding space.
% Due to this property, our loss focuses more on pairs that can contribute more on learning the target embedding space, unlike previous methods that treat samples equally without considering their relative importance~\cite{park2019relational,yu2019learning}. 
% Also, we present a simple yet effective data augmentation strategy that enables target models to consider semantic relations between multiple views of each sample as well as those between different samples.

To transfer the knowledge effectively, we propose a new loss, called \emph{relaxed contrastive loss}, that is used for learning target embedding models with the knowledge in the form of pairwise similarities. 
Our loss utilizes the pairwise similarities as \emph{relaxed labels} of inter-sample relations, unlike conventional metric learning losses that rely on binary labels indicating class equivalence between samples (\ie, 1 if two samples are of the same class and 0 otherwise) as supervision. 
By replacing the binary labels with the pairwise similarities, our loss can provide rich supervision beyond what the binary labels offer, such as the degree of similarity and hardness of a pair of training samples.

Specifically, the proposed loss pushes apart or pulls together a pair of samples in a target embedding space following the principle of the original contrastive loss~\cite{Hadsell2006}, but the semantic similarity of the pair given by the knowledge controls the strength of pushing and pulling.
% our analysis reveals that this behavior basically determines the distance of a pair in the target embedding space follows 
Also, we reveal that the loss lets more important pairs contribute more to learning the target embedding model, thus resolves the limitation of previous methods that treat samples equally during transfer~\cite{park2019relational,yu2019learning}. 
In addition to the use of relaxed relation labels, we further modify the loss so that it does not impose any restriction on the manifold of target embedding space, unlike conventional losses that enforce target embedding spaces $\ell_2$ normalized. 
This modification enables to utilize given embedding dimensions more effectively and provides extra performance improvement.
% For further improvement, we also present a data augmentation strategy, which allows to transfer semantic relations between multiple views of each sample as well as those between different samples.

The efficacy of the proposed method is first demonstrated on public benchmark datasets for deep metric learning.
Our method substantially improves image retrieval performance when the target model has the same architecture as its source counterpart, and greatly reduces the size and embedding dimension of the target model with a negligible performance drop when the target model is smaller than the source model, as shown in Fig.~\ref{fig:teaser}. 
% ---------- 20210323
% We also show that deep networks trained in a self-supervised manner~\cite{chen2020simple} can be further enhanced by self embedding transfer with our method, analogous to the born-again network~\cite{furlanello2018born} yet with no supervision. 
% In all the experiments, our method outperforms existing embedding transfer techniques~\cite{passalis2018learning,park2019relational,chen2017darkrank}.
% ---------- 20210323
We also show that our method enhances the quality of self-supervised representation through self embedding transfer and the performance of classification models in the knowledge distillation setting.
In all the experiments, our method outperforms existing embedding transfer techniques~\cite{passalis2018learning,park2019relational,chen2017darkrank}.

% \suha{The contribution of this work is three-fold:}

% \suha{The contribution of this work is three-fold:
% \begin{itemize}
%   \item a novel method for embedding transfer; capturing rich semantic info of source embedding space as knowledge and transferring transparently the knowledge to any arbitrary target embedding spaces.
%   \item enable to improve performance and to reduce model sizes and embedding dimensions with negligible performance drops in standard metric learning benchmarks.
%   \item improving performance of self-sup representation learning by self embedding transfer with no additional supervision.
% \end{itemize}
% }
% \suhac{this ``contribution'' part could be removed as it is sort of redundant.}

%%% RELATED WORK %%%%%%%%%%%%%%%%%%%%%%%%%%%%%%%%%%%%%%%%
% !TEX root = cvpr.tex

\section{Related Work}
\label{sec:relatedwork}

%% ==================================================
% \subsection{Learning Embedding Spaces}
\noindent \textbf{Deep metric learning.}
Deep metric learning is an approach to learning embedding spaces using class labels.
Previous work in this field has developed loss functions for modeling inter-sample relations based on class labels and reflecting them on the learned embedding spaces.
Contrastive loss~\cite{Chopra2005, Hadsell2006} pulls a pair of samples together if their class labels are the same and pushes them away otherwise.
Triplet loss~\cite{Wang2014, Schroff2015} takes a triplet of anchor, positive, and negative as input, and makes the anchor-positive distance smaller than the anchor-negative distance.
The idea of pushing and pulling a pair is extended to consider higher order relations in recently proposed losses~\cite{Sohn_nips2016,songCVPR16,wang2019multi}.
% Further, to solve the inherent complexity issue of methods that consider pairwise relations, methods introducing proxies, which are learnable parameters representing subsets of data are also proposed such as Proxy-NCA~\cite{movshovitz2017no} and Proxy-Anchor~\cite{kim2020proxy}. 
%
Meanwhile, self-supervised representation learning
% , which aims to learn visual representation from unlabeled data, 
has been greatly advanced by leveraging pairwise relations between data as in deep metric learning. 
For example, MoCo~\cite{he2020momentum,chen2020improved} and SimCLR~\cite{chen2020simple}
% based on contrastive learning 
pull embedding vectors of the same image closer and push those of different images away.
Since these approaches to learning embedding spaces demand binary relations, \ie, the equality of classes or identities, they cannot be used directly for transferring knowledge of an embedding space that is not binary.

\noindent\textbf{Knowledge distillation.}
Knowledge distillation means a technique that transfers knowledge of a source model to a target model; embedding transfer can be regarded as its variant focusing on metric learning.
A seminal work by Hinton \etal~\cite{hinton2015distilling} achieves this goal by encouraging the target model to imitate class logits of the source model,
% for the purpose of model compression.
and has been extended to transfer various types of knowledge of the source model~\cite{ romero2014fitnets,zagoruyko2016paying,yim2017gift,Ahn_2019_CVPR}.
Knowledge distillation has been employed for various purposes including model compression~\cite{hinton2015distilling, romero2014fitnets, zagoruyko2016paying, yim2017gift}, cross-modality learning~\cite{tian2019contrastive}, and network regularization~\cite{yun2020regularizing} as well as performance improvement~\cite{furlanello2018born}.
% In terms of target task, however, it has been applied mostly to classification and studied rarely for learning embedding spaces.
In terms of target task, however, it has been applied mostly to classification; only a few methods introduced in the next paragraph study transferring knowledge of embedding spaces, \ie, embedding transfer.

\noindent\textbf{Embedding transfer.}
% Prior to our work, there have been several attempts to distill and transfer knowledge of embedding spaces.
Early approaches in this area extract and transfer the rank of similarities between samples~\cite{chen2017darkrank} and probability distributions of their similarities~\cite{passalis2018learning} in the source embedding spaces. 
Unfortunately, these methods have trouble capturing elaborate relations between samples.
On the other hand, recently proposed methods utilize geometric relations between samples like distances and angles as the knowledge to take fine details of the source embedding space into account~\cite{park2019relational,yu2019learning}.
% However, they blindly accept the knowledge without considering relative importance of samples delivering the knowledge
% However, they let the target model blindly accept the knowledge without considering the relative importance of samples, leading to less effective embedding transfer. 
% Our method overcomes the aforementioned limitations: It makes use of rich pairwise similarities between samples as the knowledge, and the proposed loss enables to take relative importance of samples into account when transferring the knowledge.
However, they let the target model blindly accept the knowledge without considering the relative importance of samples, leading to less effective embedding transfer. 
Our method overcomes the aforementioned limitations: It makes use of rich pairwise similarities between samples as the knowledge, and can take relative importance of samples into account when transferring the knowledge.

\section{Proposed Method}
\label{sec:method}
This section first introduces the problem formulation of embedding transfer, then reviews briefly the original contrastive loss~\cite{Hadsell2006} and describes the derivation of the relaxed contrastive loss in detail.
It also discusses the effect of label relation with empirical evidences.

\subsection{Problem Formulation of Embedding Transfer}
\label{sec:embedding_transfer}

Embedding transfer is the task of transferring knowledge from a source embedding model $s$ to a target embedding model $t$. 
Let $f^s: \mathcal{X} \rightarrow \mathcal{Z}^s$ and $f^t: \mathcal{X} \rightarrow \mathcal{Z}^t$ denote the two embedding models, which are mapping functions from the same data space $\mathcal{X}$ to their own embedding spaces. 
The goal of embedding transfer is to transfer knowledge captured in $\mathcal{Z}^s$ to $\mathcal{Z}^t$ for various purposes like performance enhancement, embedding dimension reduction, and embedding model compression.
%
% Note that the dimensions of $\mathcal{Z}^s$ and $\mathcal{Z}^t$ do not have to be the same.
% $dim(\mathcal{Z}^s)>dim(\mathcal{Z}^t)$, and $dim(\mathcal{Z}^s)=dim(\mathcal{Z}^t)$
%
%
\iffalse
\sy{
Embedding transfer is a task of transferring the knowledge from the source embedding model $s$ to the target embedding model $t$. 
Given the data space $\mathcal{X}$, we let $f^s: \mathcal{X} \rightarrow \mathcal{Z}^s$ and $f^t: \mathcal{X} \rightarrow \mathcal{Z}^t$ be the source embedding model and target embedding model that map from the data space to different embedding spaces, respectively. 
Embedding transfer aims to transfer rich relational information between samples contained in the source embedding space $\mathcal{Z}^s$ to the target embedding space $\mathcal{Z}^t$, it allows to further improve performance, reduce the space dimension, or compress the embedding model.
% The objective of embedding transfer is to learn mapping from source embedding space $\mathcal{Z}^s$ to target embedding space $\mathcal{Z}^t$.
% \begin{align}
%     \mathcal{L}(X) = \sum_{{x_1, \dots , x_n}\in \mathcal{X}^N}{l(f^s(x_1, \dots, x_n),f^t(x_1,\dots, x_n))},
%     \label{eq:contrastive}
% \end{align}
}
% ==================================================
\fi

\subsection{Revisiting Original Contrastive Loss}
\label{sec:contrastive_loss}

% \cite{Hadsell2006} introduced the contrastive loss based on pairwise distance to learn discriminative embedding spaces. This loss is the most representative method for learning semantic embedding space by leveraging pairwise relation. 
Contrastive loss~\cite{Hadsell2006} is one of the most representative losses for learning semantic embedding by leveraging pairwise relations of samples.
Let $f_i := f(x_i)$ be the embedding vector of input data $x_i$ produced by the embedding network $f$, 
% and $D(f_i, f_j)$ denote the Euclidean distance between embedding vectors $f_i$ and $f_j$. 
and $d_{ij}:= ||f_i - f_j||_2$ denote the Euclidean distance between embedding vectors $f_i$ and $f_j$.
The contrastive loss is then formulated as % as input and is formulated as 
% \suhac{some of terms in this paragraph are used without appropriate introduction, e.g., positive/negative pairs and embedding vectors}
% \begin{equation}
% \ell(f_i,f_j) = y_{ij} {d(f_i, f_j)}^2 + (1-y_{ij})\Big[\delta - d(f_i, f_j)\Big]_+^2
% \label{eq:contrastive}
% \end{equation}
%
% ==================================================
% D(f_i,f_j) version
% \begin{align}
%     \begin{split}
%     \mathcal{L}(X) = & \underbrace{\frac{1}{n} \sum_{i=1}^{n}{\sum_{j=1}^{n}{y_{ij} D(f_i, f_j)^2}}}_{\text{\emph{attracting}}} \\
%     & + \underbrace{{\frac{1}{n} \sum_{i=1}^{n}{\sum_{j=1}^{n}{(1-y_{ij})\big[\delta - D(f_i, f_j)\big]_+^2}}}}_{\text{\emph{repelling}}},
%     \end{split}
%     \label{eq:contrastive}
% \end{align}
% ==================================================
% ==================================================
% D_{ij} version
\begin{align}
    \begin{split}
    \mathcal{L}(X) = \underbrace{\frac{1}{n} \sum_{i=1}^{n}{\sum_{j=1}^{n}{y_{ij} d_{ij}^2}}}_{\text{\emph{attracting}}}
    + \underbrace{{\frac{1}{n} \sum_{i=1}^{n}{\sum_{j=1}^{n}{(1-y_{ij})\big[\delta - d_{ij}\big]_+^2}}}}_{\text{\emph{repelling}}},
    \end{split}
    \label{eq:contrastive}
    % \nonumber
\end{align}
% ==================================================
%
% \suhac{would recommend to use small characters for scalar values or labels like $Y$ and $W$ in the equations since capital letters usually indicate matrices.}
where $X$ is a batch of embedding vectors, $n$ is the number of samples in the batch, $\delta$ is a margin, and $[\cdot]_+$ denotes the hinge function. 
% Also, $y_{ij}$ is $1$ indicates that the pair of samples $(i,j)$ are of the same class (i.e., positive pair), and $0$ for a pair of samples are of the different classes (i.e. negative pair) based on given binary labels. 
Also, $y_{ij}$ is the binary label indicating the class equivalence between the pair of samples $(i,j)$: $y_{ij}=1$ if the pair is of the same class (\ie, positive pair), and 0 otherwise (\ie, negative pair).
% Note that all embedding vectors should be $L_2$ normalized to prevent margin from becoming trivial. 
Note that all embedding vectors are $l_2$ normalized to prevent the margin from becoming trivial.
This loss consists of two constituents, an attracting term and a repelling term.
In the embedding space, the attracting term forces positive pairs to be closer, and the repelling term encourages to push negative pairs apart beyond the margin.

The gradient of the loss with respect to $d_{ij}$ is given by
\iffalse
\begin{equation}
    \frac{\partial \ell(f_i,f_j)}{\partial d(f_i,f_j)} =
    \begin{dcases}
    2d(f_i, f_j) , & \textrm{if } y_{ij} = 1, \\[0.3em]
    -2\Big(\delta - d(f_i, f_j)\Big), 
    & \textrm{else if } y_{ij} = 0 \textrm{ and } d(f_i,f_j) < \delta, \\[0.3em]
    0 , & \textrm{otherwise.}
    \end{dcases} 
\label{eq:contrastive_grad}
\end{equation}
\fi
%
% ==================================================
% D_{ij} version
\begin{equation}
    \frac{\partial \mathcal{L}(X)}{\partial d_{ij}} =
    \begin{dcases}
    \frac{2}{n}d_{ij} , & \textrm{if } y_{ij} = 1, \\
    \frac{2}{n}(d_{ij} - \delta), 
    & \textrm{else if } y_{ij} = 0  \textrm{ and }  d_{ij} < \delta, \\
    0 , & \textrm{otherwise.}
    \end{dcases} 
\label{eq:contrastive_grad}
\end{equation}
% ==================================================
%
As shown in Eq.~(\ref{eq:contrastive_grad}), the magnitude of the gradient increases as the distance of a positive pair increases or the distance of a negative pair decreases. When the distance of a negative pair is larger than the margin $\delta$, the gradient becomes 0. 
% As shown in Eq.~(\ref{eq:contrastive_grad}), the gradient increases as the distance of a positive pair increases or that of a negative pair decreases. When the distance of a negative pair is larger $\delta$, the gradient becomes 0.

% Note that the loss takes into account discrepancy with the goal (0 for positives, $\delta$ for negatives) of each pair \suhac{don't understand this sentence clearly}, but not semantic elaborate relations between samples.

% \sy{(Eq.2 and Eq.5 are too long to be reduced. So it seems we need to use short term of distance. (e.g. $D_{ij}, D_{ij}^s, D_{ij}^t$) What do you want to think about this?)} \suhac{agree}

\iffalse
\begin{figure} [!t]
\centering
\includegraphics[width=.85 \textwidth]{figs/figure2.pdf}
\vspace*{-2mm}
\caption{}
\label{fig:gradient_comparison}
\end{figure}
\fi

% \begin{wrapfigure}{r}{0.45\textwidth}
%   \begin{center}
%     \includegraphics[width=0.43\textwidth]{figs/fig2_small.pdf}
%   \end{center}
% %   \vspace{-1mm}
%   \caption{Gradient of the smooth contrastive loss versus pairwise distance.}
%   \label{fig:gradient_comparison}
% \end{wrapfigure}

\subsection{Relaxed Contrastive Loss}

The basic idea of the relaxed contrastive loss is to pull or push a pair of samples in the target embedding space according to their semantic similarity captured in the source embedding space as knowledge. 
% The loss is inspired by the original contrastive loss, but to reflect the knowledge, it utilizes pairwise similarities given in the knowledge as supervision that relaxes binary labels for class equivalence.
Unlike the original contrastive loss, it relaxes the binary labels indicating class equivalence relations using pairwise similarities given in the transferred knowledge.

% A straightforward way to implement this idea is to replace the class equivalence indicator $y_{ij}$ of the original contrastive loss with the semantic similarity between $x_i$ and $x_j$ in the source embedding space.
% An initial form of smooth contrastive loss can be obtained by replacing the class equivalence indicator $y_{ij}$ of the original contrastive loss with the semantic similarity between $x_i$ and $x_j$ in the source embedding space. 
This idea can be implemented simply by replacing $y_{ij}$ in Eq.~\eqref{eq:contrastive} with the semantic similarity of $x_i$ and $x_j$ in the source embedding space. 
The loss then becomes a linear combination of the attracting and repelling terms in which their weights (\ie, relaxation of $y_{ij}$) are proportional to the semantic similarities. 
Specifically, it is formulated as
\begin{align}
\begin{split}
\mathcal{L}(X) = \frac{1}{n} \sum_{i=1}^{n}{\sum_{j=1}^{n}{w_{ij}^s {d_{ij}^t}^2}} + \frac{1}{n} \sum_{i=1}^{n}{\sum_{j=1}^{n}{(1-w_{ij}^s)\Big[\delta - d_{ij}^t\Big]_+^2}},
\label{eq:smooth_contrastive}
\end{split}
\end{align}
where $w_{ij}^s$ is the weight derived from the semantic similarities in the source embedding space, 
$f_i^t := f^t(x_i) \in \mathbb{R}^d$ indicates the embedding vector of input $x_i$ produced by $f^t$,
and $d_{ij}^t$ is Euclidean distance between target embedding vectors $f_i^t$ and $f_j^t$.
For computing the weight terms, we employ a Gaussian kernel based on the Euclidean distance as follows:
% The \emph{source similarity weight} is similarity score between samples in the source embedding space and can be defined in several ways. We employ a Gaussian kernel based on the Euclidean distance as the weight and given by
%
% \begin{equation}
% w_{ij}^s = K_G(f_i^s, f_j^s; \sigma) = \exp\bigg(-\frac{{||f_i^s - f_j^s||_2^2}}{\sigma}\bigg) \in [0,1],
% \label{eq:teacher similarity}
% \end{equation}
%
\begin{equation}
w_{ij}^s = \exp\bigg(-\frac{{||f_i^s - f_j^s||_2^2}}{\sigma}\bigg) \in [0,1],
\label{eq:teacher similarity}
\end{equation}
% \suhac{use $\exp$ instead of $exp$, and bigger parentheses if the inner term is a fractional expression; see the above eq.}
% \suhac{also, as already mentioned, use small characters for scalar values like $W$.}
where $\sigma$ is kernel bandwidth
$f_i^s := f^s(x_i) $ indicates the embedding vector of input $x_i$ given by the $f^s$, 
and $||\cdot||_2$ denotes $l_2$ norm of vector.
% The use of the Gaussian kernel allows the loss to focus more on semantically similar pairs
% Note that the weights are fixed during training of the target embedding model.

Eq.~(\ref{eq:smooth_contrastive}) shows that the strength of pulling or pushing embedding vectors is now controlled by the weights in the new loss function.
In the target embedding space, a pair of samples that the source embedding model regards more similar attract each other more strongly while those considered more dissimilar are pushed more heavily out of the margin $\delta$.
This behavior of the loss can be explained through its gradient, which is given by
% As shown in Eq.~(\ref{eq:smooth_contrastive}), our loss is constructed as linear interpolations of attracting loss and repelling loss with respect to source similarity weight $w_{ij}^S$.
% In the proposed loss, the degrees of strength of pulling or pushing embedding vectors is controlled by the weight. Specifically, this loss aims that pairs that the source model considers similar are attracted more strongly to each other, and pairs that are considered dissimilar are pushed more strongly out of the margin $\delta$ in the target embedding space. Samples in the target embedding space are penalized by the sum of these two losses.
% % Unlike conventional contrastive loss, which aims to attract all positive samples as much as possible and push all negative samples beyond the margin, our method acknowledges that there is a difference in similarity between positive samples or negative samples. 
% This property of smooth contrastive loss can be explained by its gradient, which is given by
%
\iffalse
\begin{align}
\footnotesize
    \frac{\partial \mathcal{L}(X)}{\partial D(f_i^t,f_j^t)} =
    \begin{dcases}
    \frac{2}{n} \big\{ D(f_i^t, f_j^t) - \delta (1-w_{ij}^s) \big\} , & \textrm{if }  D(f_i^t,f_j^t) < \delta, \\[0.4em]
    \frac{2}{n} w_{ij}^s D(f_i^t, f_j^t), & \textrm{otherwise. }
    \end{dcases} \label{eq:smooth_contrastive_grad}
\end{align}
\fi
% ====================================================
% D_ij version
\begin{align}
    \frac{\partial \mathcal{L}(X)}{\partial d_{ij}^t} =
    \begin{dcases}
    \frac{2}{n} \big\{ d_{ij}^t - \delta (1-w_{ij}^s) \big\} , & \textrm{if }  d_{ij}^t < \delta, \\[0.4em]
    \frac{2}{n} w_{ij}^s d_{ij}^t, & \textrm{otherwise. }
    \end{dcases} \label{eq:smooth_contrastive_grad}
\end{align}
%
% Fig.~\ref{fig:gradient_comparison} qualitatively compares this gradient to that of the original contrastive loss.
%
Unlike the original one, the aspect of our loss gradient depends on the transferred knowledge $w_{ij}^s$,
thus the force of pushing a pair $(i, j)$ apart and that of pulling them together are determined by both $d_{ij}^t$ and $w_{ij}^s$.
In the ideal case, $d_{ij}^t$ will converge to $\delta (1-w_{ij}^s)$, which is the semantic dissimilarity scaled by $\delta$, and where the two forces are balanced.

This aspect of gradient also differentiates our method from the previous arts that imitate the knowledge through regression losses~\cite{park2019relational,yu2019learning}.
%
%
% \suha{As illustrated in Fig.~X(b)}, the proposed loss rarely cares about a pair $(i,j)$ whose distance is large in both of the source and target spaces, i.e., $w_{ij}^s\approx 0$ and $D(f_i^t, f_j^t) > \delta$, as the gradient of such a pair is close to 0.
% This behavior can be interpreted as that our loss disregards less important pairs so that it focuses on more important ones.
% % Note that, in the general use of learned embedding spaces, the distance of a semantically dissimilar pair matters little if it is large enough, which is a common idea in metric learning~\cite{Xing_2002,Hadsell2006,Schroff2015}.
% \suha{Note that, since semantic relations between only nearby samples matter in the general use of learned embedding spaces, the distance of a semantically dissimilar pair does not provide useful information when it is larger than a certain threshold, which is a common idea in metric learning~\cite{Xing_2002,Hadsell2006,Schroff2015}.} \suhac{will revise}
%
%
As illustrated in Fig.~\ref{fig:gradient_comparison}, the proposed loss rarely cares about a pair $(i,j)$ when its distance is large in both the source and target spaces, \ie, $w_{ij}^s\approx 0$ and $d_{ij}^t > \delta$, as its loss gradient is close to 0.
This behavior can be interpreted as that our loss disregards less important pairs to focus on more important ones.
Recall that what we expect from a learned embedding space is that \emph{nearby} samples are semantically similar in the space;
% the distance of a semantically dissimilar pair does not impair such a quality of the embedding space thus can be regarded as less important if its distance is sufficiently large.
if the distance of a semantically dissimilar pair is sufficiently large, it does not impair such a quality of the embedding space and can be regarded as less important consequently. 
% ; this is a common idea in metric learning~\cite{Xing_2002,Hadsell2006,Schroff2015}. 
On the other hand, the previous methods using regression losses handle samples equivalently without considering their importance~\cite{park2019relational,yu2019learning},
leading to suboptimal results. % and less flexible. 

The loss in Eq.~(\ref{eq:smooth_contrastive}) takes advantage of the rich semantic information of the source embedding space in a flexible and effective manner, but it still has a problem to be resolved: It imposes a restriction on the manifold of the target space since it demands $l_2$ normalization of the embedding vectors to prevent the divergence of their magnitudes and to keep the margin non-trivial, as in the original contrastive loss. 
To resolve this issue, we replace the pairwise distances of the loss in Eq.~(\ref{eq:smooth_contrastive}) with their \emph{relative} versions, then the final form of the relaxed contrastive loss is given by
%
\iffalse
\begin{align}
\begin{split}
\mathcal{L}(X) =& \frac{1}{n} \sum_{i=1}^{n}{\sum_{j=1}^{n}{w_{ij}^s \bigg\{\frac{D(f_i^t, f_j^t)}{\mu_i}\bigg\}^2}} \\  &+ \frac{1}{n} \sum_{i=1}^{n}{\sum_{j=1}^{n}{(1-w_{ij}^s)\bigg[\delta - \frac{D(f_i^t, f_j^t)}{\mu_i}\bigg]_+^2}}, \\
&\qquad\:\textrm{where } \mu_i = \frac{1}{n} \sum_{k=1}^{n}{D(f_i^t,f_k^t)}. 
\label{eq:smooth_contrastive_mu}
\end{split}
\end{align}
\fi
% ====================================================
% D_ij version
\begin{align}
\begin{split}
\mathcal{L}(X) =& \frac{1}{n} \sum_{i=1}^{n}{\sum_{j=1}^{n}{w_{ij}^s \bigg(\frac{d_{ij}^t}{\mu_i}\bigg)^2}} \\
&+ \frac{1}{n} \sum_{i=1}^{n}{\sum_{j=1}^{n}{(1-w_{ij}^s)\bigg[\delta - \frac{d_{ij}^t}{\mu_i}\bigg]_+^2}}, \\
&\quad\textrm{where } 
\mu_i = \frac{1}{n} \sum_{k=1}^{n}{d_{ij}^t}. 
\label{eq:smooth_contrastive_mu}
\end{split}
\end{align}
The relative distance between $f_i^t$ and $f_j^t$ is their pairwise distance divided by $\mu_i$, the average distance of all pairs associated with $f_i^t$ in the batch.
Since scales of pairwise distances are roughly canceled in their relative versions, the above loss can alleviate the aforementioned normalization issue.
Thus, although the source embedding space is limited to the surface of unit hypersphere due to the $l_2$ normalization, the target embedding model can exploit the entire space of $\mathbb{R}^d$ with no restriction on its manifold.
% This advantage enables to utilize the given embedding dimensions more effectively.
We found empirically that this advantage helps improve performance of target embedding models and reduce their embedding dimensions effectively.

\begin{figure} [!t]
\centering
\includegraphics[width = 0.85 \columnwidth]{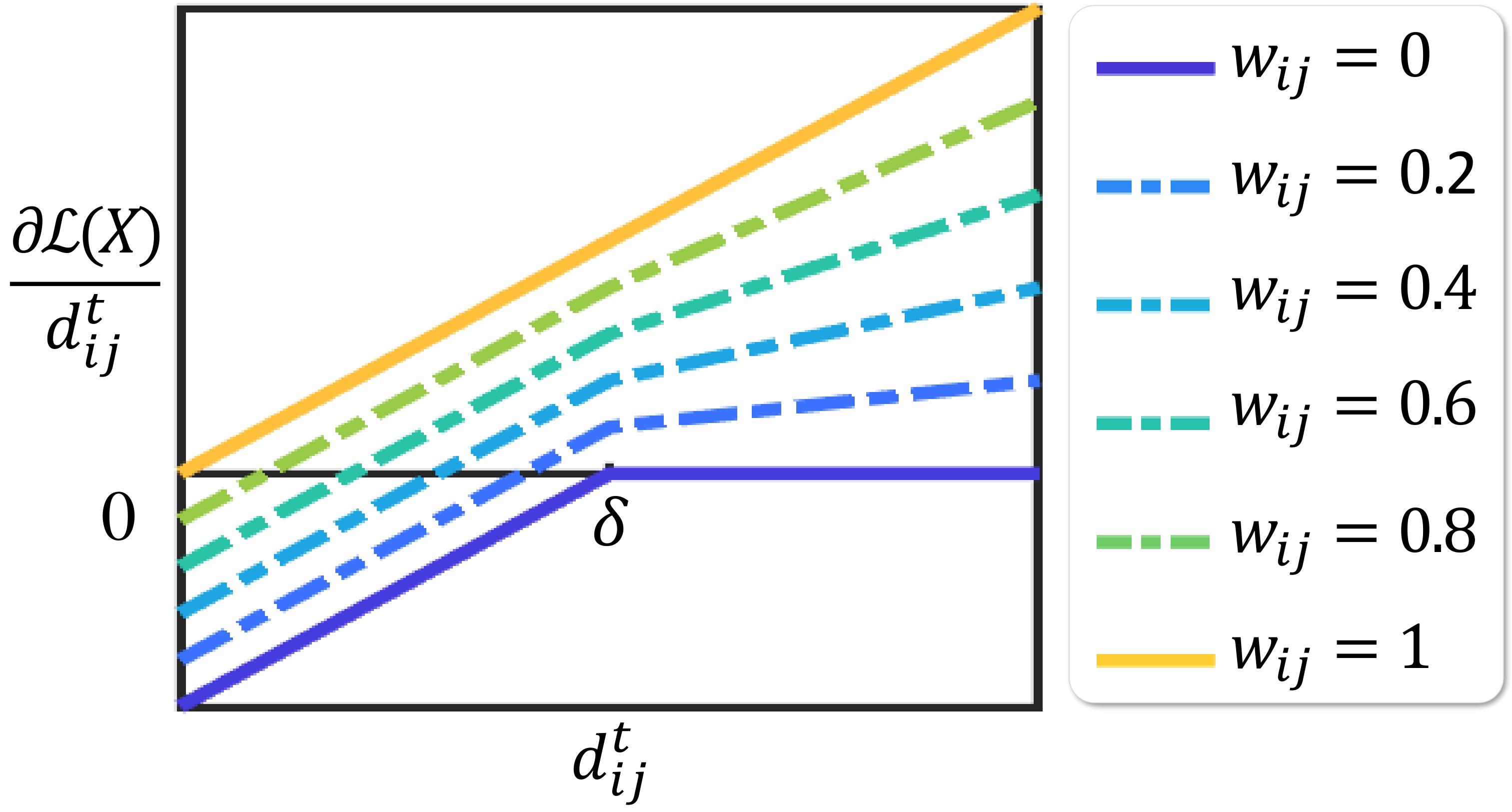}
\caption{Gradients of relaxed contrastive loss versus pairwise distance given different weights.}
\label{fig:gradient_comparison}
\vspace*{-1mm}
\end{figure}

\subsection{Discussion on Label Relaxation}
\label{sec:appendix_knowledge}

\begin{figure*} [!t]
\centering
\includegraphics[width = 0.98 \textwidth]{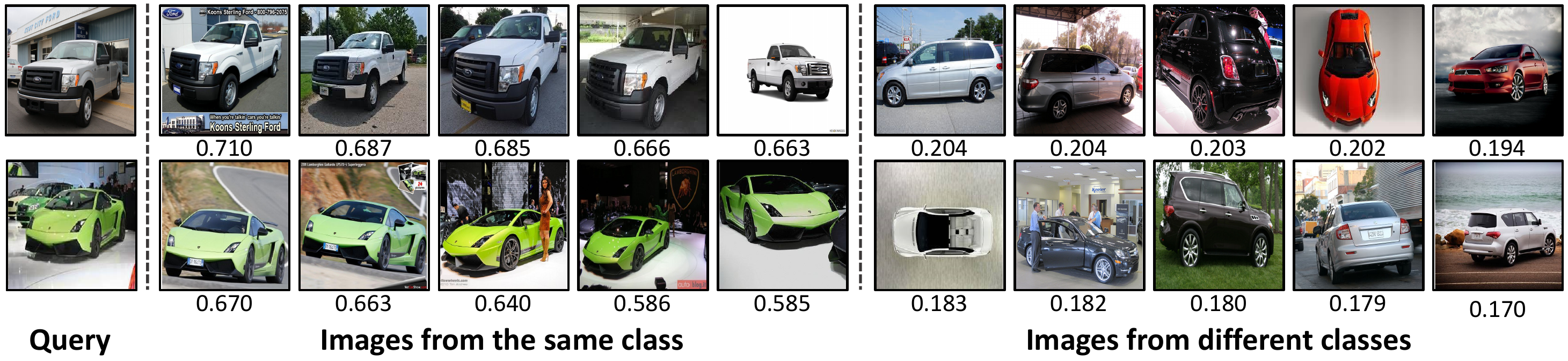}
\caption{
% Qualitative results according to extracted weights on CUB-200-2011 dataset.
Image pairs sorted by their normalized weights of Eq.~(\ref{eq:teacher similarity}) on the Cars-196 dataset. 
% The leftmost images are paired with the other images in their rows. 
}
\label{fig:weights_cars196}
\vspace*{-2mm}
\end{figure*}

\begin{figure} [!t]
\centering
\includegraphics[width = 0.97 \columnwidth]{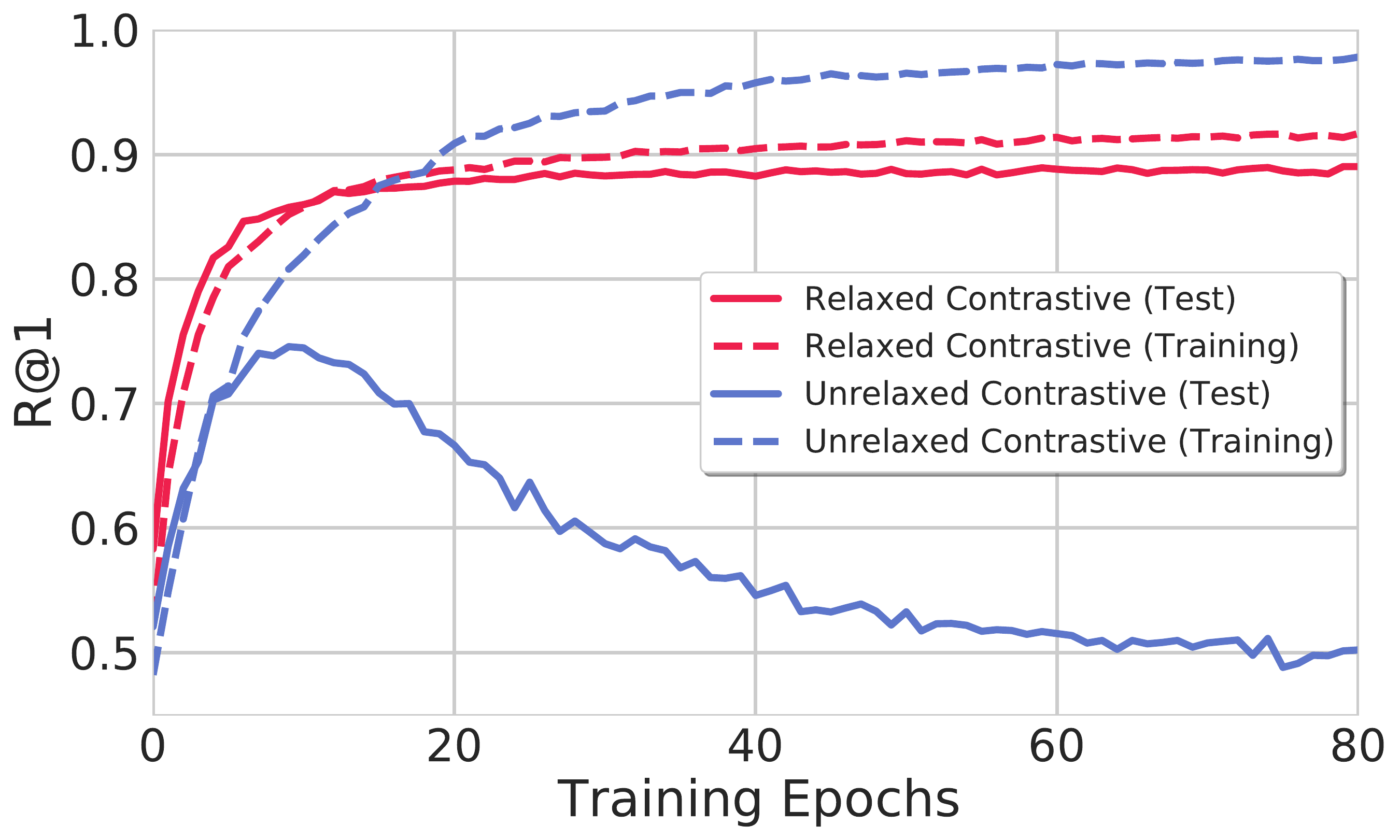}
\caption{Accuracy in Recall@1 versus epochs on the Cars-196 dataset~\cite{krause20133d}.
The dotted and solid lines represent training and test accuracy, respectively.
}
\label{fig:accuracy_relax_unrelax}
\vspace*{-1mm}
\end{figure}

Label relaxation allows our loss to exploit rich information such as degree of similarity between samples, within-class variation, and between-class affinity, all of which cannot be offered by the binary inter-sample relations. 
To demonstrate this property empirically, we in Fig.~\ref{fig:weights_cars196} enumerate image pairs with top-5 and bottom-5  normalized weights, \ie, $w^s_{ij}$ of Eq.~\eqref{eq:teacher similarity}. 
As shown in the figure, pairs exhibiting more similar poses or backgrounds have higher weights even in the same class 
while those of different classes showing large appearance variations are assigned low weights.

% Label relaxation thus improves generalization of target models by providing such rich and diverse supervisory signals, in contrast to the binary labels that allow learning how to discriminate different classes and lead to degraded performance on unseen classes consequently.
Label relaxation thus improves generalization of target models by providing such rich and diverse supervisory signals, in contrast to the binary labels which only allow the model to learn to discriminate different classes and lead to degraded performance on unseen classes consequently.
% We argue that the main advantage of label relaxation is improved generalization of target embedding models. 
This is demonstrated by evaluating two embedding models trained by the relaxed contrastive loss and its unrelaxed version using $y_{ij}$ instead of $w^s_{ij}$ in Eq.~(\ref{eq:smooth_contrastive_mu}), respectively.
% To demonstrate this effect, we evaluate two embedding models trained by the relaxed contrastive loss and its unrelaxed version using $y_{ij}$ instead of $w^s_{ij}$ in Eq.~(\ref{eq:smooth_contrastive_mu}), respectively.
% ; the variant \emph{binary contrastive loss}.
% as it is identical to the original contrastive loss except that the pairwise distances are computed in the relative manner.
Fig.~\ref{fig:accuracy_relax_unrelax} compares the performance of the two models on the training and test splits of the Cars-196 dataset~\cite{krause20133d}.
As shown in the figure, relaxed contrastive loss helps the model generalize well to unseen test data while the model trained by the unrelaxed version is quickly overfitted to training data. 
% \suhac{why and how does it improve generalization?}

Our label relaxation method is independent of loss functions, thus can be integrated with other metric learning losses based on pairwise relations of samples.
% We yet choose relaxed contrastive loss as our loss due to its simplicity, interpretability, and superior performance.
Relaxed contrastive loss is yet chosen as our loss due to its simplicity, interpretability, and superior performance.
% The reasons for choosing contrastive loss are its simplicity, interpretability, and superior performance. 
We have also applied the same method to Multi-Similarity (MS) loss~\cite{wang2019multi}, and observed that relaxed MS loss achieves comparable performance but demands more hyper-parameters and careful tuning of them.
More analysis and comparisons are given in the supplementary material.

\section{Experiments}
\label{sec:experiments}
This section demonstrates the effectiveness of embedding transfer by our method in three different tasks, deep metric learning, self-supervised representation learning, and knowledge distillation for classification. 

\begin{table*}[!t]
\centering
\small
\begin{tabularx}{\textwidth}{ 
   >{\centering\arraybackslash}X |
   >{\centering\arraybackslash}X |
   >{\centering\arraybackslash}X |
   >{\centering\arraybackslash}X 
   >{\centering\arraybackslash}X
   >{\centering\arraybackslash}X |
   >{\centering\arraybackslash}X
   >{\centering\arraybackslash}X 
   >{\centering\arraybackslash}X |
   >{\centering\arraybackslash}X
   >{\centering\arraybackslash}X
   >{\centering\arraybackslash}X  }
\hline
\multicolumn{3}{l|}{\multirow{2}{*}[-2mm]{Recall@$K$}}         & \multicolumn{3}{c|}{CUB-200-2011} & \multicolumn{3}{c|}{Cars-196} & \multicolumn{3}{c}{SOP}  \\ 
\cline{4-12}
\multicolumn{3}{l|}{}  & 1 & 2 & 4  & 1 & 2 & 4 & 1 & 10 & 100             \\ 
\hline
\multirow{8}{*}{(a)}& \multicolumn{1}{l|}{\emph{Source}: PA~\cite{kim2020proxy}} & \multicolumn{1}{l|}{BN$^{512}$} &69.1 &78.9 &86.1 &86.4 &91.9 &95.0 &79.2 &90.7 &96.2 \\
& \multicolumn{1}{l|}{FitNet~\cite{romero2014fitnets}} & \multicolumn{1}{l|}{BN$^{512}$} &69.9& 79.5& 86.2& 87.6& 92.2& 95.6& \underline{78.7}& \underline{90.4}&96.1  \\
& \multicolumn{1}{l|}{Attention~\cite{zagoruyko2016paying}} & \multicolumn{1}{l|}{BN$^{512}$} & 66.3& 76.2& 84.5& 84.7& 90.6& 94.2& 78.2& 90.4&\underline{96.2} \\
& \multicolumn{1}{l|}{CRD~\cite{tian2019contrastive}} & \multicolumn{1}{l|}{BN$^{512}$} & 67.7& 78.1& 85.7& 85.3& 91.1& 94.8& 78.1& 90.2&95.8  \\
& \multicolumn{1}{l|}{DarkRank~\cite{chen2017darkrank}} & \multicolumn{1}{l|}{BN$^{512}$} &66.7 &76.5 &84.8 &84.0 &90.0 &93.8 &75.7 &88.3 &95.3 \\
& \multicolumn{1}{l|}{PKT~\cite{passalis2018learning}}  & \multicolumn{1}{l|}{BN$^{512}$} &69.1 &78.8 &86.4 &86.4 &91.6 &94.9 &78.4 &90.2 &96.0 \\
& \multicolumn{1}{l|}{RKD ~\cite{park2019relational}}  & \multicolumn{1}{l|}{BN$^{512}$} & \underline{70.9} & \underline{80.8} &\underline{87.5} &\underline{88.9} &\underline{93.5} & \underline{96.4} &78.5 &90.2 &96.0   \\
% & \multicolumn{1}{l|}{\ccol Ours w/o augmentation} & \multicolumn{1}{l|}{\ccol BN$^{512}$} &\ccol \underline{71.5} &\ccol \underline{81.3} &\ccol \underline{87.6} &\ccol \underline{89.0} &\ccol \underline{93.6} &\ccol 96.3 & \ccol \underline{79.6} &\ccol \underline{91.0}&\ccol \underline{96.2} \\
& \multicolumn{1}{l|}{\ccol Ours} & \multicolumn{1}{l|}{\ccol BN$^{512}$} &\ccol \textbf{72.1} &\ccol \textbf{81.3} &\ccol \textbf{87.6} &\ccol \textbf{89.6} &\ccol\textbf{94.0} &\ccol \textbf{96.5}  &\ccol \textbf{79.8} &\ccol \textbf{91.1} &\ccol\textbf{96.3} \\ 
\hline

\multirow{8}{*}{(b)} & \multicolumn{1}{l|}{\emph{Source}: PA~\cite{kim2020proxy}} &\multicolumn{1}{l|}{BN$^{512}$} &69.1 &78.9 &86.1 &86.4 &91.9 &95.0 &79.2 &90.7 &96.2 \\ 
& \multicolumn{1}{l|}{FitNet~\cite{romero2014fitnets}} & \multicolumn{1}{l|}{BN$^{64}$} & 62.3& 73.8& 83.0& 81.2& 87.7& 92.5& \textbf{76.6}& \textbf{89.3}&\textbf{95.4} \\
& \multicolumn{1}{l|}{Attention~\cite{zagoruyko2016paying}} & \multicolumn{1}{l|}{BN$^{64}$} & 58.3& 69.4& 79.1&79.2& 86.7& 91.8& 76.3& \underline{89.2} & \underline{95.4} \\
& \multicolumn{1}{l|}{CRD~\cite{tian2019contrastive}} & \multicolumn{1}{l|}{BN$^{64}$} & 60.9& 72.7& 81.7& 79.2& 87.2& 92.1 & 75.5& 88.3&95.3 \\
& \multicolumn{1}{l|}{DarkRank~\cite{chen2017darkrank}}& \multicolumn{1}{l|}{BN$^{64}$} &63.5 &74.3 &83.1 &78.1 &85.9 &91.1 &73.9 &{87.5} &{94.8} \\
& \multicolumn{1}{l|}{PKT~\cite{passalis2018learning}}  & \multicolumn{1}{l|}{BN$^{64}$} &63.6 &75.8 &84.0 &82.2 &88.7 &93.5 &74.6 &87.3 &94.2 \\
& \multicolumn{1}{l|}{RKD~\cite{park2019relational}}  & \multicolumn{1}{l|}{BN$^{64}$}  &\underline{65.8} &\underline{76.7} &\underline{85.0} &\underline{83.7} &\underline{89.9} &\underline{94.1} &70.2 &83.8 &92.1 \\ 
% & \multicolumn{1}{l|}{\ccol Ours w/o augmentation} &\multicolumn{1}{l|}{\ccol BN$^{64}$}  &\ccol \underline{67.1} &\ccol \underline{77.8} &\ccol \underline{85.8} &\ccol \underline{85.2}&\ccol \underline{91.0}&\ccol \underline{95.0}&\ccol \underline{75.7}&\ccol 86.2&\ccol \underline{94.7}\\
& \multicolumn{1}{l|}{\ccol Ours} & \multicolumn{1}{l|}{\ccol BN$^{64}$} &\ccol \textbf{67.4}&\ccol \textbf{78.0}&\ccol \textbf{85.9}&\ccol \textbf{86.5}&\ccol \textbf{92.3}&\ccol \textbf{95.3}&\ccol \underline{76.3}&\ccol {88.6}&\ccol 94.8 \\ 
\hline

\multirow{8}{*}{(c)} & \multicolumn{1}{l|}{\emph{Source}: PA~\cite{kim2020proxy}}  & \multicolumn{1}{c|}{R50$^{512}$} & 69.9& 79.6& 88.6& 87.7& 92.7& 95.5& 80.5& 91.8& 98.8   \\
& \multicolumn{1}{l|}{FitNet~\cite{romero2014fitnets}} & \multicolumn{1}{l|}{R18$^{128}$} & 61.0& 72.2& 81.1& 78.5& 86.0& 91.4& 76.7&\underline{89.4} &95.5 \\
& \multicolumn{1}{l|}{Attention~\cite{zagoruyko2016paying}} & \multicolumn{1}{l|}{R18$^{128}$} & 61.0& 71.7& 81.5& 78.6&85.9 &91.0 & 76.4&89.3 &95.5 \\
& \multicolumn{1}{l|}{CRD~\cite{tian2019contrastive}} & \multicolumn{1}{l|}{R18$^{128}$} & 62.8& 73.8& 83.2& 80.6& 87.9& 92.5& 76.2& 88.9&95.3  \\
& \multicolumn{1}{l|}{DarkRank~\cite{chen2017darkrank}} & \multicolumn{1}{l|}{R18$^{128}$} &61.2 &72.5 &82.0 &75.3 &83.6 &89.4 &72.7 &86.7 &94.5  \\
& \multicolumn{1}{l|}{PKT~\cite{passalis2018learning}}   & \multicolumn{1}{c|}{R18$^{128}$} &65.0 &75.6 &84.8 &81.6 &88.8 &93.4 &\underline{76.9} &89.2 &\underline{95.5}  \\
& \multicolumn{1}{l|}{RKD~\cite{park2019relational}}  & \multicolumn{1}{l|}{R18$^{128}$} &\underline{65.8} &\underline{76.3} &\underline{84.8} &\underline{84.2} &\underline{90.4} &\underline{94.3} &75.7 &88.4 &95.1   \\ 
% & \multicolumn{1}{l|}{\ccol Ours w/o augmentation} & \multicolumn{1}{l|}{\ccol R18$^{128}$} &\ccol \underline{66.4}&\ccol \underline{77.4}&\ccol \underline{85.3}&\ccol \underline{84.5}&\ccol \underline{91.0}&\ccol \underline{94.9}&\ccol \underline{77.8}&\ccol \underline{90.0}&\ccol \underline{95.8}\\
& \multicolumn{1}{l|}{\ccol Ours}  & \multicolumn{1}{l|}{\ccol R18$^{128}$} &\ccol \textbf{66.6}&\ccol \textbf{78.1}&\ccol \textbf{85.9}&\ccol \textbf{86.0}&\ccol \textbf{91.6}&\ccol \textbf{95.3}&\ccol \textbf{78.4}&\ccol \textbf{90.4}&\ccol \textbf{96.1}\\ 
\hline
\end{tabularx}
\vspace{0.1mm}
\caption{Image retrieval performance of embedding transfer and knowledge distillation methods in the three different settings: (a) Self-transfer, (b) dimensionality reduction, and (c) model compression.  
Embedding networks of the methods are denoted by abbreviations: BN--Inception with BatchNorm, R50--ResNet50, R18--ResNet18.
Superscripts indicate embedding dimensions of the networks.}
\label{tab:emb_transfer_comp}
\vspace{-1mm}
\end{table*}

%% ==================================================
\subsection{Deep Metric Learning}

On standard benchmarks for metric learning, we evaluate and compare target embedding models trained by embedding transfer methods, RKD~\cite{park2019relational},  PKT~\cite{passalis2018learning}, and DarkRank~\cite{chen2017darkrank} as well as ours.
These methods train target models solely by embedding transfer losses; no other supervision is required for the target model.
% In the embedding transfer methods, target models are trained solely by embedding transfer losses; no other supervision is introduced for the models. 
In addition, three knowledge distillation techniques, FitNet~\cite{romero2014fitnets}, Attention~\cite{zagoruyko2016paying}, and CRD~\cite{tian2019contrastive} are also evaluated on the same datasets to examine their effectiveness for embedding transfer.\footnote{Knowledge distillation methods relying on classification logits cannot be applied to our task where source model has no classification layer.} 
In this case, knowledge distillation losses are coupled with a metric learning loss since they extract knowledge from intermediate layers of the source model and are not directly aware of its embedding space consequently.

The experiments are conducted in the following three settings by varying the type of target model.
\emph{(i) Self-transfer for performance improvement}: Transfer to a model with the same architecture and embedding dimension.
\emph{(ii) Dimensionality reduction}: Transfer to the same architecture with a lower embedding dimension.
\emph{(iii) Model compression}: Transfer to a smaller network with a lower embedding dimension.

\subsubsection{Setup}
\noindent \textbf{Datasets and evaluation.} 
Target models are evaluated in terms of image retrieval performance on the CUB-200-2011~\cite{CUB200}, Cars-196~\cite{krause20133d} and SOP datasets~\cite{songCVPR16}. 
Each dataset is split into training and test sets following the standard setting presented in~\cite{songCVPR16}.
As a performance measure, we adopt Recall@$K$ that counts how many queries have at least
one correct sample among their $K$ nearest neighbors in learned embedding spaces.

% We evaluated the image retrieval performance on CUB200-2011~\cite{CUB200}, Cars196 ~\cite{krause20133d} and Stanford Online Products datasets~\cite{songCVPR16}. We split the three dataset for training and testing, following the standard data split setting of the metric learning as in~\cite{songCVPR16}.~\sungyeon{We can add on more details about datasets.}

\noindent \textbf{Source and target embedding networks.} 
For the \emph{self-transfer} and \emph{dimensionality reduction} experiments, we employ BatchNorm Inception~\cite{Batchnorm} with 512 output dimension as the source model. 
Target models for the two settings basically have the same architecture as the source model, but for \emph{dimensionality reduction}, the output dimension is reduced to 64. 
On the other hand, in the \emph{model compression} experiment, we adopt ResNet50~\cite{resnet} with 512 output dimension as the source model and ResNet18~\cite{resnet} with 128 output dimension as the target model. 
In all the three settings, the source models are trained by proxy-anchor loss~\cite{kim2020proxy} with $l_2$ normalization of embedding vectors, while the target models are pre-trained for the ImageNet classification task~\cite{Imagenet} and have no $l_2$ normalization applied.

\noindent \textbf{Implementation details.} 
We train all the models using the AdamW optimizer~\cite{adamw} with the cosine learning decay~\cite{loshchilov2016sgdr} and initial learning rate of $10^{-4}$. 
They are learned for 90 epochs in the CUB-200-2011 and Cars-196 datasets, and 150 epochs on the SOP dataset.
Training images are randomly flipped horizontally and cropped to $224 \times 224$, and test images are center-cropped after being resized to $256 \times 256$. 
Further, we generate two different views of each image in a batch by the random augmentations; details and effects of this augmentation strategy are described in the supplementary material.
% The multi-view augmentation strategy applies random augmentation operation twice per image so that input batch contains two different views of each image, \sy{and the details of the augmentation strategy are described in the supplementary material.}
We set both $\delta$ and $\sigma$ in our loss to 1 for all the experiments. 
For knowledge distillation, Proxy-Anchor loss~\cite{kim2020proxy} is coupled with distillation losses using the same weight.

\subsubsection{Results}
% The proposed method is evaluated and compared with existing embedding transfer techniques and state-of-the-art metric learning models on the three benchmark datasets. 
% We also report the performance of our method without the multi-view data augmentation strategy to demonstrate its effectiveness.
% The proposed method is evaluated and compared with existing knowledge distillation techniques including embedding transfer approaches and state-of-the-art metric learning models on the three benchmark datasets. The results are summarized in Table~\ref{tab:emb_transfer_comp} and~\ref{tab:dml_cub_cars_sop}.
The proposed method is compared to the embedding transfer and knowledge distillation methods in terms of performance of target embedding models on the three benchmark datasets in Table~\ref{tab:emb_transfer_comp}.
Its records are also compared with those of state-of-the-art metric learning methods on the same datasets in Table~\ref{tab:dml_cub_cars_sop}.

\begin{figure*} [!t]
\centering
\includegraphics[width = 1 \textwidth]{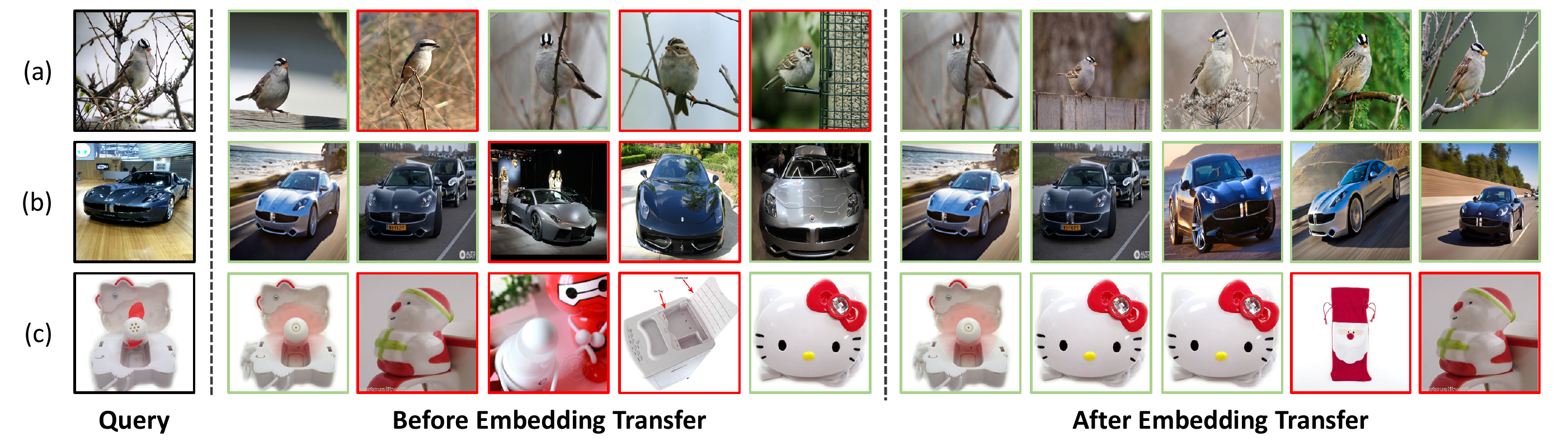}
\caption{
% Qualitative examples of t
Top 5 image retrievals of the state of the art~\cite{kim2020proxy} before and after the proposed method is applied.
(a) CUB-2020-2011. (b) Cars-196. (c) SOP.
Images with green boundary are success cases and those with red boundary are false positives.
More qualitative results can be found in the supplementary material.}
\label{fig:qualitative_results}
\vspace{-2mm}
\end{figure*}

In the \emph{self-transfer} setting  (Table~\ref{tab:emb_transfer_comp}(a)), the proposed method notably improves retrieval performance and clearly surpasses the state of the art
on all the datasets without bells and whistles (Table~\ref{tab:dml_cub_cars_sop}); 
the effect of embedding transfer by our method is qualitatively demonstrated in Fig.~\ref{fig:qualitative_results}. 
On the other hand, the performance of existing embedding transfer methods is inferior to that of the source model on the SOP dataset. 
% Consequently, our method clearly outperforms the state-of-the-art metric learning model and other embedding transfer techniques in this setting. 
The proposed method demonstrates more interesting results in the \emph{dimensionality reduction} setting  (Table~\ref{tab:emb_transfer_comp}(b)): It outperforms recent metric learning methods, MS and DiVA,
% ~\cite{wang2019multi,milbich2020diva} 
whose embedding dimension is 8 times higher (Table~\ref{tab:dml_cub_cars_sop}).
This result enables significant speedup of image retrieval systems at the cost of a tiny performance drop.
Finally, in the \emph{model compression} setting (Table~\ref{tab:emb_transfer_comp}(c)), our method achieves impressive performance even with a substantially smaller network and a lower embedding dimension; the performance drop by the compression is marginal and its accuracy is as competitive as MS with a heavier network and a larger embedding dimension.
% Note that, the proposed method is superior to other embedding transfer techniques in every experiment, and the multi-view augmentation strategy improves performance in most cases.

% Conventional knowledge distillation methods tend to underperform embedding transfer methods on datasets excluding SOP. Since these distillation techniques cannot directly transfer the knowledge of the source embedding space, their performance are highly dependent on the that of metric learning loss applied together. In contrast, the proposed method is superior to the existing knowledge distillation method in most experiments without relying on task loss and using additional memory like CRD.
We found that the knowledge distillation methods tend to underperform, especially on the CUB-200-2011 and Cars-196 datasets.
In particular, their performance depends heavily on the coupled metric learning loss since they cannot directly transfer knowledge of the source embedding space. 
In contrast, our method is superior to them in most experiments with no additional loss nor memory buffer~\cite{tian2019contrastive}.

\begin{table*}[!t]
\centering
\small
\begin{tabularx}{\textwidth}{ 
   >{\centering\arraybackslash}X
   >{\centering\arraybackslash}X |
   >{\centering\arraybackslash}X 
   >{\centering\arraybackslash}X
   >{\centering\arraybackslash}X |
   >{\centering\arraybackslash}X
   >{\centering\arraybackslash}X 
   >{\centering\arraybackslash}X |
   >{\centering\arraybackslash}X
   >{\centering\arraybackslash}X
   >{\centering\arraybackslash}X  }
 \hline
\multicolumn{2}{l|}{\multirow{2}{*}[-2mm]{Recall@$K$}} & \multicolumn{3}{c|}{CUB-200-2011} & \multicolumn{3}{c|}{Cars-196} & \multicolumn{3}{c}{SOP}\\ \cline{3-11}
\multicolumn{2}{l|}{} &1 &2 &4 &1 &2 &4 &1 &10 &100 \\ \hline
\multicolumn{1}{l|}{MS~\cite{wang2019multi}}& \multicolumn{1}{l|}{BN$^{512}$}& 65.7 &77.0 & \underline{86.3} &84.1 &90.4 &94.0 &78.2 &90.5 &96.0  \\
\multicolumn{1}{l|}{SoftTriple~\cite{Qian_2019_ICCV}}& \multicolumn{1}{l|}{BN$^{512}$}& 65.4 & 76.4 & 84.5& 84.5 & 90.7 & 94.5 & 78.3& 90.3& 95.9 \\
\multicolumn{1}{l|}{DiVA~\cite{milbich2020diva}}& \multicolumn{1}{l|}{BN$^{512}$}& 66.8& 77.7& -& 84.1& 90.7& -&78.1 &90.6 & -  \\
\multicolumn{1}{l|}{PA~\cite{kim2020proxy}}& \multicolumn{1}{l|}{BN$^{512}$}& \underline{69.1} & \underline{78.9} & 86.1 & \underline{86.4} & \underline{91.9} & \underline{95.0} & \underline{79.2} & \underline{90.7} & \underline{96.2}\\
\multicolumn{1}{l|}{\ccol Ours} & \multicolumn{1}{l|}{\ccol BN$^{512}$} &\ccol \textbf{72.1} &\ccol \textbf{81.3} &\ccol \textbf{87.6} &\ccol \textbf{89.6} &\ccol\textbf{94.0} &\ccol \textbf{96.5}  &\ccol \textbf{79.8} &\ccol \textbf{91.1} &\ccol\textbf{96.3} \\ 
\hline

\multicolumn{1}{l|}{MS~\cite{wang2019multi}}& \multicolumn{1}{l|}{BN$^{64}$}& 57.4& 69.8& 80.0& 77.3& 85.3& 90.5& 74.1& 87.8& 94.7 \\
\multicolumn{1}{l|}{SoftTriple~\cite{Qian_2019_ICCV}}& \multicolumn{1}{l|}{BN$^{64}$}& 60.1& 71.9& 81.2& 78.6& 86.6 & 91.8 & \underline{76.3}& \textbf{89.1}&  \textbf{95.3} \\
\multicolumn{1}{l|}{DiVA~\cite{milbich2020diva}}& \multicolumn{1}{l|}{BN$^{64}$}& \underline{63.0} & \underline{74.5}& \underline{83.3} & 78.3& 86.6& 91.2 & 73.7 & 87.5& 94.8\\
\multicolumn{1}{l|}{PA~\cite{kim2020proxy}}& \multicolumn{1}{l|}{BN$^{64}$}& 61.7& 73.0& 81.8& \underline{78.8}& \underline{87.0}& \underline{92.2}& \textbf{76.5}& \underline{89.0}& \underline{95.1} \\
\multicolumn{1}{l|}{\ccol Ours} & \multicolumn{1}{l|}{\ccol BN$^{64}$} &\ccol \textbf{67.4}&\ccol \textbf{78.0}&\ccol \textbf{85.9}&\ccol \textbf{86.5}&\ccol \textbf{92.3}&\ccol \textbf{95.3}&\ccol {76.3} &\ccol {88.6} &\ccol 94.8 \\ 
\hline

\multicolumn{1}{l|}{PA~\cite{kim2020proxy}}& \multicolumn{1}{l|}{R18$^{128}$}& \underline{61.8}& \underline{72.9}& \underline{82.1}& \underline{78.7}& \underline{86.5}& \underline{91.7}& \underline{76.2}& \underline{89.1}& \underline{95.2} \\
\multicolumn{1}{l|}{\ccol Ours} & \multicolumn{1}{l|}{\ccol R18$^{128}$} &\ccol \textbf{66.6}&\ccol \textbf{78.1}&\ccol \textbf{85.9}&\ccol \textbf{86.0}&\ccol \textbf{91.6}&\ccol \textbf{95.3}&\ccol \textbf{78.4} &\ccol \textbf{90.4} &\ccol \textbf{96.1} \\ 
\hline

\end{tabularx}
\vspace{0.1mm}
\caption{
Image retrieval performance of the proposed method and the state-of-the-art metric learning models. Embedding networks of the methods are fixed by Inception with BatchNorm (BN) for fair comparisons, and superscripts indicate embedding dimensions.
}
\label{tab:dml_cub_cars_sop}
% \vspace{-1mm}
\end{table*}

\subsubsection{Ablation Study}
We conduct ablation study in the \emph{self-transfer} setting to examine the contribution of each component of the relaxed contrastive loss.
 % \sy{compared to the original contrastive loss using binary label as supervision}. 
The results are summarized in Table~\ref{tab:ablation}.

We first validate the effect of label relaxation by replacing indicator $y_{ij}$ in Eq.~\eqref{eq:contrastive} with semantic similarity $w_{ij}^s$ obtained from source embedding model. 
The result suggests that label relaxation significantly improves performance by exploiting the rich semantic relations between samples.
Also, we investigate the effect of using the relative distance, \ie, removing $l_2$ normalization to be free from the restriction on the embedding manifold.
The result shows that adopting the relative distance further improves the performance as it allows to fully exploit the entire embedding dimensions with no restriction; a similar observation has been reported in~\cite{park2019relational}.

\begin{table}[!t]
\centering
\small
\begin{tabularx}{0.48 \textwidth}{ 
   >{\centering\arraybackslash}X |
   >{\centering\arraybackslash}X |
   >{\centering\arraybackslash}X} 
\hline
\multicolumn{1}{l|}{\multirow{2}{*}[0mm]{Methods}} & \multicolumn{2}{c}{Recall@1} \\  \cline{2-3}
\multicolumn{1}{l|}{} & CUB& Cars\\ \hline
 \multicolumn{1}{l|}{Original contrastive loss} & 65.3 & 80.5\\ 
 \multicolumn{1}{l|}{+ Label relaxation} &  70.4& 88.2\\ 
 % \multicolumn{1}{l|}{+ Removing $l_2$ norm} & 72.1 & 89.6\\ \hline
 \multicolumn{1}{l|}{+ Relative distance} & 72.1 & 89.6\\ \hline
%  \multicolumn{1}{l|}{Hard thresholding} & 65.1 & 77.7\\ \hline
%  \multicolumn{1}{l|}{Ours w/o augmentation} &  71.5 & 89.0\\ 
\end{tabularx}
\vspace{0.1mm}
\caption{
% Comparison of source embedding and ablations of our method on the CUB-200-2011 (CUB) and Cars-196 (Cars) datasets.
Ablation study of the components of our loss on the CUB-200-2011 (CUB) and Cars-196 (Cars) datasets. 
}
\label{tab:ablation}
\vspace{-1mm}
\end{table}

\subsection{Self-supervised Representation Learning}

Knowledge distillation has been known to improve the performance of classification models by self-transfer~\cite{furlanello2018born}, but is not available for self-supervised representation learning due to the absence of class labels.
We argue that embedding transfer can play this role for networks trained in a self-supervised manner since it distills and transfers knowledge without relying on class labels.
% , in the way that knowledge extracted from self-supervised network

This section examines a potential of embedding transfer methods in this context, by learning representations using the embedding transfer methods and the knowledge extracted from existing self-supervised networks.
Our method is compared with RKD~\cite{park2019relational} and PKT~\cite{passalis2018learning}, but DarkRank~\cite{chen2017darkrank} is excluded since its complexity, proportional to the number of sample permutations, is excessively large in the self-supervised learning setting.
We adopt a network trained by SimCLR~\cite{chen2020simple}, the state of the art in self-supervised representation learning, as the source embedding model.

\subsubsection{Setup}
\textbf{Datasets and evaluation.}
% The representation of the model trained with embedding transfer on CIFAR-10~\cite{krizhevsky2009learning} and STL-10~\cite{coates2011analysis} is evaluated by linear evaluation, which is a commonly used for evaluation in self-supervised learning. In the experiment of STL-10, labeled training set and unlabeled set are used for training, and the rest set for testing. Following the linear evaluation protocol~\cite{zhang2016colorful, bachman2019learning, zhai2019s4l}, we measure the test accuracy after training linear classifier on the top of the frozen base network.
Self-supervised models and those enhanced by embedding transfer are evaluated on the CIFAR-10~\cite{krizhevsky2009learning} and STL-10~\cite{coates2011analysis} datasets.
In the STL-10 dataset, both the labeled training set and unlabeled set are used for training, and the rest are kept for testing.
The performance of the models is measured by the linear evaluation protocol~\cite{zhang2016colorful, bachman2019learning, zhai2019s4l}, in which a linear classifier on top of a frozen self-supervised network is trained and evaluated.

\iffalse
We use CIFAR-10~\cite{krizhevsky2009learning} and STL-10~\cite{coates2011analysis} for the evaluation of this task. Since STL-10 is divided into a labeled training set and an unlabeled set, we use all these sets during training. We evaluate the representation of the model trained with our method in linear evaluation, which is a commonly used for evaluation in self-supervised learning. Following the procedure described in~\cite{zhang2016colorful, bachman2019learning, zhai2019s4l}, we measure the test accuracy after training linear classifier on the top of the frozen base network. 
\fi
% We freeze all the parameters of the base network of the trained model and append a linear classifier. Then, we measure the test accuracy after training only the parameters of classifier using labels.

% \textbf{Training settings for source model.}
% For this experiment, we reimplement SimCLR~\cite{chen2020simple} and MoCo v2~\cite{chen2020improved} and use for pretraining source embedding models. Following~\cite{chen2020simple}, we employ ResNet-50 for base network of source models, and some layers are changed to handle small image sizes. Depending on the setting of each method, the MLP head are appended to the last pooling layer of models. We trained the source embedding model for 1000 epochs on CIFAR-10 according to the settings described in~\cite{chen2020simple} and published code (e.g, augmentation, learning rate and temperature).
% In STL-10, the same settings as in CIFAR-10 are used, except that Gaussian blur is additionally used for augmentation.

\noindent \textbf{Source models and their training.}
We reimplement SimCLR~\cite{chen2020simple} framework to train source embedding models. 
% For the base network of source models, we employ ResNet50 and append the MLP head to the last pooling layer of the model following~\cite{chen2020simple}. 
Following the original framework, ResNet50 is employed as the base network of the source models and a Multi-Layer Perceptron (MLP) head is appended to its last pooling layer.
% Following~\cite{chen2020simple}, ResNet50 is employed as the based network of source models, and a Multi-Layer Perceptron (MLP) head is appended to the last pooling layer of the network.
On the CIFAR-10 dataset, the source models are trained for 1K epochs while following the details (\eg, augmentation, learning rate, and temperature) described in \cite{chen2020simple}.
On the STL-10 dataset, we adopt the same configuration except that Gaussian blur is additionally employed for data augmentation.

\noindent \textbf{Target models and their training.}
For training of target models, the MLP on top of the source models are removed and their embedding vectors are $l_2$ normalized.
% so that the target models are trained with normalized embedding vectors. 
Target models have the same architecture as their source counterpart where the MLP head is removed, but with no $l_2$ normalization.
Details of training target models are the same as those for the corresponding source models. 
All target models are trained using the LARS optimizer~\cite{you2017large} with initial learning rate of $4.0$ and weight decay of $10^{-6}$.
We warm up the learning rate linearly during the first 10 epochs 
% and decay it using the cosine learning decay schedule~\cite{loshchilov2016sgdr}.
and apply the cosine decay~\cite{loshchilov2016sgdr} after that.
Regarding hyperparameters, both $\delta$ and $\sigma$ in our loss are set to 1.

\subsubsection{Results}
The performance of embedding transfer methods in the self-supervised learning task is summarized in Table~\ref{tab:eval_cifar_stl}. 
The proposed method improves the quality of the learned representations on both of the two datasets.
 % and for both self-supervised learning frameworks.
Moreover, it clearly outperforms the existing embedding transfer techniques when incorporated with SimCLR. 
In contrast, other embedding transfer methods are often inferior to the source model, and especially PKT shows unstable performance in every experiment.
This is because of their limitations: As the batch size increases, the probability distributions considered by PKT becomes nearly uniform, and the computational burden of RKD grows significantly due to its angle calculation.
Our method enhances the performance of the existing self-supervised models without such difficulties.

\begin{table}
\centering
\small
\begin{tabularx}{0.48\textwidth}{ 
   >{\centering\arraybackslash}X |
   >{\centering\arraybackslash}X |
   >{\centering\arraybackslash}X } 
\hline
\multicolumn{1}{l|}{\multirow{1}{*}{Dataset}} & CIFAR-10 &  STL-10  \\ 
\cline{2-3} \hline
 \multicolumn{1}{l|}{\emph{Before embedding transfer}} &  93.4 & 89.2\\ %\hline
 \multicolumn{1}{l|}{PKT~\cite{passalis2018learning}} &  65.3 & 71.6\\ 
 \multicolumn{1}{l|}{RKD~\cite{park2019relational}} &  \underline{93.6} &  \underline{79.8} \\ 
 \multicolumn{1}{l|}{\ccol Ours} & \ccol \textbf{93.9} & \ccol \textbf{89.6}\\ \hline
\end{tabularx}
\vspace{0.1mm}
\caption{Performance of linear classifiers trained on representations obtained by embedding transfer techniques incorporated with self-supervised learning frameworks.}
\label{tab:eval_cifar_stl}
\vspace{-1mm}
\end{table}

% Classification KD
\begin{table}[!t]
\centering
\small
\begin{tabularx}{0.48\textwidth}{ 
   >{\centering\arraybackslash}X |
   >{\centering\arraybackslash}X |
   >{\centering\arraybackslash}X} 
\hline
\multicolumn{1}{l|}{Source}  & ResNet56 (72.34)& VGG13 (74.64)\\ 
\multicolumn{1}{l|}{Target}  & ResNet20 (69.06) & VGG8 (70.36)\\ \hline
% \multicolumn{1}{c|}{Source} &  & \\
% \multicolumn{1}{c|}{Target} & 69.06 & \\ \hline
\multicolumn{1}{l|}{HKD~\cite{hinton2015distilling}} & 70.66 & 72.98\\ 
\multicolumn{1}{l|}{RKD~\cite{park2019relational} + HKD} & 71.18 & 72.97\\ 
\multicolumn{1}{l|}{CRD~\cite{tian2019contrastive} + HKD} & \underline{71.63} & \textbf{74.29}\\
\multicolumn{1}{l|}{\ccol Ours + HKD} & \ccol \textbf{71.95} & \ccol \underline{73.82}\\  \hline
\end{tabularx}
\vspace{0.1mm}
\caption{
Test accuracy of target models on the CIFAR100 dataset.
% of knowledge distillation methods. 
}
\label{tab:KD_CIFAR100}
\vspace{-1mm}
\end{table}

%% ==================================================
\subsection{Image Classification}
% \sy{
% Finally, we compare the proposed method with several state-of-the-art knowledge distillation methods to prove that it is also effective in image classification. We perform experiments in the model compression setting on the CIFAR-100 dataset, and adopted the source-target model combination of ResNet56-ResNet20 and VGG13-VGG8. All methods including ours is applied cross-entropy loss additionally and combined with HKD~\cite{hinton2015distilling}, which is applied on the classification layer of source and target models. Our loss uses the embedding vector generated from the last pooling layer of each model and that of the source model is $l_2$ normalized. We directly follow the training detail described in \cite{tian2019contrastive} and both $\delta$ and $\sigma$ in our loss are set to 1.
% }

Finally, we demonstrate that the proposed method can be used also for enhancing classifiers as a knowledge distillation technique.
Following the convention in this task, its efficacy is validated in the model compression setting on the CIFAR-100~\cite{krizhevsky2009learning} dataset with two source--target combinations: ResNet56--ResNet20~\cite{resnet} and VGG13--VGG8~\cite{vggnet}.
Our method is compared with RKD~\cite{hinton2015distilling} and CRD~\cite{tian2019contrastive} as well as HKD~\cite{hinton2015distilling};
all methods including ours are combined with HKD and use the cross-entropy loss additionally. 
In detail, our relaxed contrastive loss utilizes the outputs from the last pooling layer of the source and target models, and the embedding vectors of the source are $l_2$ normalized.
We directly follow \cite{tian2019contrastive} for other training details, and both $\delta$ and $\sigma$ in our loss are set to 1.

\iffalse
% To investigate the scalability of the proposed method, we evaluate and compare our loss with RKD~\cite{hinton2015distilling}, CRD~\cite{tian2019contrastive} and HKD~\cite{hinton2015distilling} on the image classification task.
To investigate the efficacy of the proposed method on the image classification task, we compare the proposed method with RKD~\cite{hinton2015distilling}, CRD~\cite{tian2019contrastive} and HKD~\cite{hinton2015distilling}.
As the general evaluation protocol for KD methods, we perform experiments in the model compression setting on the CIFAR-100~\cite{krizhevsky2009learning} dataset, and adopt two source-target combinations, \ie, ResNet56-ResNet20~\cite{resnet} and VGG13-VGG8~\cite{vggnet}.
All methods including ours are combined with HKD and use cross-entropy loss additionally. 
% Our loss utilizes the embedding vectors produced from the last pooling layer of the source and target models, and the former's $l_2$ normalized.
% The embedding vectors utilized in our loss are produced from the last pooling layer of the source and target models, and those of the former are $l_2$ normalized.
Our loss utilizes the outputs produced from the last pooling layer of the source and target models, and the embedding vectors of the source are $l_2$ normalized.
We directly follow the training details described in \cite{tian2019contrastive} and both $\delta$ and $\sigma$ in our loss are set to 1.
\fi

As shown in Table~\ref{tab:KD_CIFAR100}, our method is comparable to or outperforming the state of the art~\cite{tian2019contrastive}.
This result indicates that our method is universal and can be applied to tasks other than metric learning.
% Our method was applied to KD for classification and compared with existing KD methods on the CIFAR100 dataset; all methods including ours are integrated with HKD~\cite{hinton2015distilling}.
% As shown in Table~\ref{tab:KD_CIFAR100}, our method was comparable to or outperformed state of the art; this result suggests that it is universal and can be applied to tasks other than metric learning. 

% \begin{table}[!t]
% \caption{}
% \vspace{1mm}
% % \centering
% \footnotesize
% \begin{tabularx}{\textwidth}{ 
%   >{\raggedright\arraybackslash}X |
%   >{\centering\arraybackslash}X |
%   >{\centering\arraybackslash}X }
%  \hline
%  & CIFAR-10&  STL-10 \\ \hline\hline
%  \multicolumn{3}{l}{\textbf{ResNet50}} \\ \hline
% %  SimCLR & 94.0 & \\
% %  SimCLR & & \\
%  $\dagger$SimCLR~\cite{chen2020simple} & 93.40 & 89.20\\ 
%  $\dagger$MoCo~\cite{he2020momentum} & 86.79& \\ \hline \hline
%  \multicolumn{3}{l}{\textbf{SimCLR: ResNet50 $\xrightarrow{}$ ResNet50}} \\ \hline
%  PKT \cite{passalis2018learning} & 65.27 & 78.94\\
%  RKD \cite{park2019relational} & 93.63\\ 
%  \ccol Ours &\ccol 93.92 &\ccol 89.54 \\ \hline \hline
%  \multicolumn{3}{l}{\textbf{MoCo: ResNet50 $\xrightarrow{}$ ResNet50}} \\ \hline
%  PKT \cite{passalis2018learning} & 47.64& \\
%  RKD \cite{park2019relational} & 87.72& \\ 
%  \ccol Ours &\ccol 87.65&\ccol \\ \hline 
% \end{tabularx}
% \label{tab:eval_cifar_stl}
% % \vspace*{-4mm}
% \end{table}

%%% CONCLUSION %%%%%%%%%%%%%%%%%%%%%%%%%%%%%%%%%%%%%%%%%%
% !TEX root = cvpr.tex

\section{Conclusion}
\label{sec:conclusion}
We have presented a novel method to distill and transfer knowledge of a learned embedding model effectively.
Our loss utilizes rich pairwise relations between samples in the source embedding space as the knowledge through relaxed relation labels, and effectively transfers the knowledge by focusing more on sample pairs important for learning target embedding models.
As a result, our method has achieved impressive performance over the state of the art on metric learning benchmarks and demonstrated that it can reduce the size and embedding dimension of an embedding model significantly with a negligible performance drop. 
Moreover, we have shown that our method can enhance the quality of self-supervised representation and performance of classification models.
% self-supervised representation by self embedding transfer. 

\vspace{3mm}
{
\noindent \textbf{Acknowledgement:} 
This work was supported by 
the NRF grant, % 중견, ERC
the IITP grant, % ETRI
and R\&D program for Advanced Integrated-intelligence for IDentification, % KIST
funded by Ministry of Science and ICT, Korea
(No.2019-0-01906 Artificial Intelligence Graduate School Program--POSTECH,
 NRF-2021R1A2C3012728--30\%, % 중견과제
 NRF-2018R1A5A1060031--20\%, % ERC
 NRF-2018M3E3A1057306--30\%, % KIST
 IITP-2020-0-00842--20\%). % KIST

% This work was supported by IITP grant, Basic Science Research Program, and R\&D program for Advanced Integrated-intelligence for IDentification through the NRF funded by the Ministry of Science, ICT (No.2019-0-01906 Artificial Intelligence Graduate School Program (POSTECH), NRF-2018R1C1B6001223, NRF-2018R1A5A1060031, NRF-2018M3E3A1057306, NRF-2017R1E1A1A01077999).
}

% In the future, we will explore the extension of our method that provide a dynamic curriculum tailored to the target model, rather than selecting important samples on a fixed basis.

% This work was supported by the National Research Foundation of Korea(NRF) grant funded by the Korea government(MSIT) (No. 한국연구재단에서 부여한 과제 관리번호). 
% * MSIT : Ministry of Science and ICT(과학기술정보통신부)

% This work was supported by the Institute of Information & communications Technology Planning & Evaluation(IITP) grant funded by the Korea government(MSIT) (No.2020-0-00842, Development of Cloud Robot Intelligence for Continual Adaptation to User Reactions in Real Service Environments)

{\small
\bibliographystyle{ieee_fullname}
\bibliography{cvpr21}

\begin{thebibliography}{10}\itemsep=-1pt

\bibitem{Ahn_2019_CVPR}
Sungsoo Ahn, Shell~Xu Hu, Andreas Damianou, Neil~D. Lawrence, and Zhenwen Dai.
\newblock Variational information distillation for knowledge transfer.
\newblock In {\em Proc. IEEE Conference on Computer Vision and Pattern
  Recognition (CVPR)}, 2019.

\bibitem{bachman2019learning}
Philip Bachman, R~Devon Hjelm, and William Buchwalter.
\newblock Learning representations by maximizing mutual information across
  views.
\newblock In {\em Proc. Neural Information Processing Systems (NeurIPS)}, 2019.

\bibitem{Bucher_ECCV_2016}
Maxime Bucher, St{\'{e}}phane Herbin, and Fr{\'{e}}d{\'{e}}ric Jurie.
\newblock Improving semantic embedding consistency by metric learning for
  zero-shot classification.
\newblock In {\em Proc. European Conference on Computer Vision (ECCV)}, 2016.

\bibitem{chen2020simple}
Ting Chen, Simon Kornblith, Mohammad Norouzi, and Geoffrey Hinton.
\newblock A simple framework for contrastive learning of visual
  representations.
\newblock In {\em Proc. International Conference on Machine Learning (ICML)},
  2020.

\bibitem{chen2020improved}
Xinlei Chen, Haoqi Fan, Ross Girshick, and Kaiming He.
\newblock Improved baselines with momentum contrastive learning.
\newblock {\em arXiv preprint arXiv:2003.04297}, 2020.

\bibitem{chen2017darkrank}
Yuntao Chen, Naiyan Wang, and Zhaoxiang Zhang.
\newblock Darkrank: Accelerating deep metric learning via cross sample
  similarities transfer.
\newblock In {\em Proc. AAAI Conference on Artificial Intelligence (AAAI)},
  2017.

\bibitem{Chopra2005}
S. Chopra, R. Hadsell, and Y. LeCun.
\newblock Learning a similarity metric discriminatively, with application to
  face verification.
\newblock In {\em Proc. IEEE Conference on Computer Vision and Pattern
  Recognition (CVPR)}, 2005.

\bibitem{coates2011analysis}
Adam Coates, Andrew Ng, and Honglak Lee.
\newblock An analysis of single-layer networks in unsupervised feature
  learning.
\newblock In {\em Proc. International Conference on Artificial Intelligence and
  Statistics (AISTATS)}, 2011.

\bibitem{Cox08_MDS}
Michael A.~A. Cox and Trevor~F. Cox.
\newblock {\em Multidimensional Scaling}, pages 315--347.
\newblock Springer Berlin Heidelberg, Berlin, Heidelberg, 2008.

\bibitem{Imagenet}
Jia Deng, Wei Dong, Richard Socher, Li-Jia Li, Kai Li, and Li Fei-Fei.
\newblock {ImageNet:} a large-scale hierarchical image database.
\newblock In {\em Proc. IEEE Conference on Computer Vision and Pattern
  Recognition (CVPR)}, 2009.

\bibitem{furlanello2018born}
Tommaso Furlanello, Zachary Lipton, Michael Tschannen, Laurent Itti, and Anima
  Anandkumar.
\newblock Born again neural networks.
\newblock In {\em Proc. International Conference on Machine Learning (ICML)},
  2018.

\bibitem{Hadsell2006}
R. Hadsell, S. Chopra, and Y. LeCun.
\newblock Dimensionality reduction by learning an invariant mapping.
\newblock In {\em Proc. IEEE Conference on Computer Vision and Pattern
  Recognition (CVPR)}, 2006.

\bibitem{Harwood_2017_ICCV}
Ben Harwood, Vijay Kumar B~G, Gustavo Carneiro, Ian Reid, and Tom Drummond.
\newblock Smart mining for deep metric learning.
\newblock In {\em Proc. IEEE International Conference on Computer Vision
  (ICCV)}, 2017.

\bibitem{he2020momentum}
Kaiming He, Haoqi Fan, Yuxin Wu, Saining Xie, and Ross Girshick.
\newblock Momentum contrast for unsupervised visual representation learning.
\newblock In {\em Proc. IEEE Conference on Computer Vision and Pattern
  Recognition (CVPR)}, 2020.

\bibitem{resnet}
Kaiming He, Xiangyu Zhang, Shaoqing Ren, and Jian Sun.
\newblock Deep residual learning for image recognition.
\newblock In {\em Proc. IEEE Conference on Computer Vision and Pattern
  Recognition (CVPR)}, June 2016.

\bibitem{hinton2015distilling}
Geoffrey Hinton, Oriol Vinyals, and Jeff Dean.
\newblock Distilling the knowledge in a neural network.
\newblock {\em arXiv preprint arXiv:1503.02531}, 2015.

\bibitem{Batchnorm}
Sergey Ioffe and Christian Szegedy.
\newblock Batch normalization: Accelerating deep network training by reducing
  internal covariate shift.
\newblock In {\em Proc. International Conference on Machine Learning (ICML)},
  2015.

\bibitem{JACOB_2019_ICCV}
Pierre Jacob, David Picard, Aymeric Histace, and Edouard Klein.
\newblock Metric learning with horde: High-order regularizer for deep
  embeddings.
\newblock In {\em Proc. IEEE International Conference on Computer Vision
  (ICCV)}, 2019.

\bibitem{kim2020proxy}
Sungyeon Kim, Dongwon Kim, Minsu Cho, and Suha Kwak.
\newblock Proxy anchor loss for deep metric learning.
\newblock In {\em Proc. IEEE Conference on Computer Vision and Pattern
  Recognition (CVPR)}, 2020.

\bibitem{kim2019deep}
Sungyeon Kim, Minkyo Seo, Ivan Laptev, Minsu Cho, and Suha Kwak.
\newblock Deep metric learning beyond binary supervision.
\newblock In {\em Proc. IEEE Conference on Computer Vision and Pattern
  Recognition (CVPR)}, 2019.

\bibitem{ensemble_embedding}
Wonsik Kim, Bhavya Goyal, Kunal Chawla, Jungmin Lee, and Keunjoo Kwon.
\newblock Attention-based ensemble for deep metric learning.
\newblock In {\em Proc. European Conference on Computer Vision (ECCV)}, 2018.

\bibitem{ko2020embedding}
Byungsoo Ko and Geonmo Gu.
\newblock Embedding expansion: Augmentation in embedding space for deep metric
  learning.
\newblock In {\em Proc. IEEE Conference on Computer Vision and Pattern
  Recognition (CVPR)}, 2020.

\bibitem{krause20133d}
Jonathan Krause, Michael Stark, Jia Deng, and Li Fei-Fei.
\newblock 3d object representations for fine-grained categorization.
\newblock In {\em Proceedings of the IEEE International Conference on Computer
  Vision Workshops}, pages 554--561, 2013.

\bibitem{krizhevsky2009learning}
Alex Krizhevsky and Geoffrey Hinton.
\newblock Learning multiple layers of features from tiny images.
\newblock 2009.

\bibitem{loshchilov2016sgdr}
Ilya Loshchilov and Frank Hutter.
\newblock Sgdr: Stochastic gradient descent with warm restarts.
\newblock {\em arXiv preprint arXiv:1608.03983}, 2016.

\bibitem{adamw}
Ilya Loshchilov and Frank Hutter.
\newblock Decoupled weight decay regularization.
\newblock In {\em Proc. International Conference on Learning Representations
  (ICLR)}, 2019.

\bibitem{milbich2020diva}
Timo Milbich, Karsten Roth, Homanga Bharadhwaj, Samarth Sinha, Yoshua Bengio,
  Bj{\"o}rn Ommer, and Joseph~Paul Cohen.
\newblock Diva: Diverse visual feature aggregation fordeep metric learning.
\newblock In {\em Proc. European Conference on Computer Vision (ECCV)}, 2020.

\bibitem{mohan2020moving}
Deen~Dayal Mohan, Nishant Sankaran, Dennis Fedorishin, Srirangaraj Setlur, and
  Venu Govindaraju.
\newblock Moving in the right direction: A regularization for deep metric
  learning.
\newblock In {\em Proc. IEEE Conference on Computer Vision and Pattern
  Recognition (CVPR)}, 2020.

\bibitem{movshovitz2017no}
Yair Movshovitz-Attias, Alexander Toshev, Thomas~K Leung, Sergey Ioffe, and
  Saurabh Singh.
\newblock No fuss distance metric learning using proxies.
\newblock In {\em Proc. IEEE International Conference on Computer Vision
  (ICCV)}, 2017.

\bibitem{Opitz_ICCV_2017}
M. Opitz, G. Waltner, H. Possegger, and H. Bischof.
\newblock Bier — boosting independent embeddings robustly.
\newblock In {\em Proc. IEEE International Conference on Computer Vision
  (ICCV)}, 2017.

\bibitem{opitz2018deep}
Michael Opitz, Georg Waltner, Horst Possegger, and Horst Bischof.
\newblock Deep metric learning with bier: Boosting independent embeddings
  robustly.
\newblock {\em IEEE Transactions on Pattern Analysis and Machine Intelligence
  (TPAMI)}, 2018.

\bibitem{park2019relational}
Wonpyo Park, Dongju Kim, Yan Lu, and Minsu Cho.
\newblock Relational knowledge distillation.
\newblock In {\em Proc. IEEE Conference on Computer Vision and Pattern
  Recognition (CVPR)}, 2019.

\bibitem{passalis2018learning}
Nikolaos Passalis and Anastasios Tefas.
\newblock Learning deep representations with probabilistic knowledge transfer.
\newblock In {\em Proc. European Conference on Computer Vision (ECCV)}, 2018.

\bibitem{Qian_2019_ICCV}
Qi Qian, Lei Shang, Baigui Sun, Juhua Hu, Hao Li, and Rong Jin.
\newblock Softtriple loss: Deep metric learning without triplet sampling.
\newblock In {\em Proc. IEEE International Conference on Computer Vision
  (ICCV)}, 2019.

\bibitem{Qiao_2019_ICCV}
Limeng Qiao, Yemin Shi, Jia Li, Yaowei Wang, Tiejun Huang, and Yonghong Tian.
\newblock Transductive episodic-wise adaptive metric for few-shot learning.
\newblock In {\em Proc. IEEE International Conference on Computer Vision
  (ICCV)}, 2019.

\bibitem{romero2014fitnets}
Adriana Romero, Nicolas Ballas, Samira~Ebrahimi Kahou, Antoine Chassang, Carlo
  Gatta, and Yoshua Bengio.
\newblock Fitnets: Hints for thin deep nets.
\newblock In {\em Proc. International Conference on Learning Representations
  (ICLR)}, 2014.

\bibitem{roth2020revisiting}
Karsten Roth, Timo Milbich, Samarth Sinha, Prateek Gupta, Bjoern Ommer, and
  Joseph~Paul Cohen.
\newblock Revisiting training strategies and generalization performance in deep
  metric learning.
\newblock In {\em Proc. International Conference on Machine Learning (ICML)},
  2020.

\bibitem{Schroff2015}
Florian Schroff, Dmitry Kalenichenko, and James Philbin.
\newblock {FaceNet: A unified embedding for face recognition and clustering}.
\newblock In {\em Proc. IEEE Conference on Computer Vision and Pattern
  Recognition (CVPR)}, 2015.

\bibitem{vggnet}
Karen Simonyan and Andrew Zisserman.
\newblock Very deep convolutional networks for large-scale image recognition.
\newblock In {\em Proc. International Conference on Learning Representations
  (ICLR)}, 2015.

\bibitem{snell2017prototypical}
Jake Snell, Kevin Swersky, and Richard Zemel.
\newblock Prototypical networks for few-shot learning.
\newblock In {\em Proc. Neural Information Processing Systems (NeurIPS)}, 2017.

\bibitem{Sohn_nips2016}
Kihyuk Sohn.
\newblock Improved deep metric learning with multi-class n-pair loss objective.
\newblock In {\em Proc. Neural Information Processing Systems (NeurIPS)}, 2016.

\bibitem{songCVPR16}
Hyun~Oh Song, Yu Xiang, Stefanie Jegelka, and Silvio Savarese.
\newblock Deep metric learning via lifted structured feature embedding.
\newblock In {\em Proc. IEEE Conference on Computer Vision and Pattern
  Recognition (CVPR)}, 2016.

\bibitem{sung2018learning}
Flood Sung, Yongxin Yang, Li Zhang, Tao Xiang, Philip~HS Torr, and Timothy~M
  Hospedales.
\newblock Learning to compare: Relation network for few-shot learning.
\newblock In {\em Proc. IEEE Conference on Computer Vision and Pattern
  Recognition (CVPR)}, 2018.

\bibitem{Tenenbaum00_Isomap}
Joshua~B. Tenenbaum, Vin~de Silva, and John~C. Langford.
\newblock A global geometric framework for nonlinear dimensionality reduction.
\newblock {\em Science}, 290(5500):2319--2323, 2000.

\bibitem{tian2019contrastive}
Yonglong Tian, Dilip Krishnan, and Phillip Isola.
\newblock Contrastive representation distillation.
\newblock In {\em Proc. International Conference on Learning Representations
  (ICLR)}, 2019.

\bibitem{Wang2014}
Jiang Wang, Yang Song, T. Leung, C. Rosenberg, Jingbin Wang, J. Philbin, Bo
  Chen, and Ying Wu.
\newblock Learning fine-grained image similarity with deep ranking.
\newblock In {\em Proc. IEEE Conference on Computer Vision and Pattern
  Recognition (CVPR)}, 2014.

\bibitem{wang2019multi}
Xun Wang, Xintong Han, Weilin Huang, Dengke Dong, and Matthew~R Scott.
\newblock Multi-similarity loss with general pair weighting for deep metric
  learning.
\newblock In {\em Proc. IEEE Conference on Computer Vision and Pattern
  Recognition (CVPR)}, 2019.

\bibitem{wang2020cross}
Xun Wang, Haozhi Zhang, Weilin Huang, and Matthew~R Scott.
\newblock Cross-batch memory for embedding learning.
\newblock In {\em Proc. IEEE Conference on Computer Vision and Pattern
  Recognition (CVPR)}, pages 6388--6397, 2020.

\bibitem{CUB200}
P. Welinder, S. Branson, T. Mita, C. Wah, F. Schroff, S. Belongie, and P.
  Perona.
\newblock {Caltech-UCSD Birds 200}.
\newblock Technical Report CNS-TR-2010-001, California Institute of Technology,
  2010.

\bibitem{sampling_matters}
Chao-Yuan Wu, R. Manmatha, Alexander~J. Smola, and Philipp Krahenbuhl.
\newblock Sampling matters in deep embedding learning.
\newblock In {\em Proc. IEEE International Conference on Computer Vision
  (ICCV)}, 2017.

\bibitem{yim2017gift}
Junho Yim, Donggyu Joo, Jihoon Bae, and Junmo Kim.
\newblock A gift from knowledge distillation: Fast optimization, network
  minimization and transfer learning.
\newblock In {\em Proc. IEEE Conference on Computer Vision and Pattern
  Recognition (CVPR)}, 2017.

\bibitem{you2017large}
Yang You, Igor Gitman, and Boris Ginsburg.
\newblock Large batch training of convolutional networks.
\newblock {\em arXiv preprint arXiv:1708.03888}, 2017.

\bibitem{Yu_2019_ICCV}
Baosheng Yu and Dacheng Tao.
\newblock Deep metric learning with tuplet margin loss.
\newblock In {\em Proc. IEEE International Conference on Computer Vision
  (ICCV)}, 2019.

\bibitem{yu2019learning}
Lu Yu, Vacit~Oguz Yazici, Xialei Liu, Joost van~de Weijer, Yongmei Cheng, and
  Arnau Ramisa.
\newblock Learning metrics from teachers: Compact networks for image embedding.
\newblock In {\em Proc. IEEE Conference on Computer Vision and Pattern
  Recognition (CVPR)}, 2019.

\bibitem{yun2020regularizing}
Sukmin Yun, Jongjin Park, Kimin Lee, and Jinwoo Shin.
\newblock Regularizing class-wise predictions via self-knowledge distillation.
\newblock In {\em Proc. IEEE Conference on Computer Vision and Pattern
  Recognition (CVPR)}, 2020.

\bibitem{zagoruyko2016paying}
Sergey Zagoruyko and Nikos Komodakis.
\newblock Paying more attention to attention: Improving the performance of
  convolutional neural networks via attention transfer.
\newblock In {\em Proc. International Conference on Learning Representations
  (ICLR)}, 2016.

\bibitem{zhai2019s4l}
Xiaohua Zhai, Avital Oliver, Alexander Kolesnikov, and Lucas Beyer.
\newblock S4l: Self-supervised semi-supervised learning.
\newblock In {\em Proc. IEEE International Conference on Computer Vision
  (ICCV)}, 2019.

\bibitem{zhang2016colorful}
Richard Zhang, Phillip Isola, and Alexei~A Efros.
\newblock Colorful image colorization.
\newblock In {\em Proc. European Conference on Computer Vision (ECCV)}, 2016.

\bibitem{zhang2016zero}
Ziming Zhang and Venkatesh Saligrama.
\newblock Zero-shot learning via joint latent similarity embedding.
\newblock In {\em Proc. IEEE Conference on Computer Vision and Pattern
  Recognition (CVPR)}, 2016.

\end{thebibliography}
}

%%% SUPPLEMENTARY %%%%%%%%%%%%%%%%%%%%%%%%%%%%%%%%%%%%%%%
\renewcommand\thesection{\Alph{section}}
\setcounter{section}{0}

\newpage
\section{Appendix}
This supplementary material presents deeper analyses of the proposed method, its implementation details, and additional experimental results, all of which are omitted from the main paper due to the space limit. 
% This supplementary material presents another combination of label relaxation and metric learning loss, a deeper analysis on the generalization capability of the proposed method, its implementation details, and additional experimental results, all of which are omitted from the main paper due to the space limit. 
Section~\ref{sec:relaxed_MS} first introduces relaxed MS loss, an integration of the label relaxation and Multi-Similarity (MS) loss~\cite{wang2019multi}.
The generalization capability of our model is then illustrated in terms of the spectral decay metric~\cite{roth2020revisiting} in Section~\ref{sec:generalization}.
% proposed in \cite{roth2020revisiting}.
Section~\ref{sec:appendix_augmentation} describes details of the multi-view data augmentation strategy.
In Section~\ref{sec:hyperparameters}, we investigate the effect of hyperpameters on the performance of our loss.
Finally, in Section~\ref{sec:appendix_qualitative}, we present more qualitative examples for image retrieval before and after applying the proposed method on the three metric learning benchmarks.

% This supplementary material presents the other use and a deeper analysis of the proposed method, its implementation details, and experimental results omitted from the main paper due to the space limit. 
% First, Section~\ref{sec:relaxed_MS} introduces relaxed MS loss, a formulation that applied label relaxation to Multi-Similarity (MS) loss~\cite{wang2019multi}, and Section~\ref{sec:generalization} analyzes the generalization of our model using the spectral decay.
% % proposed in \cite{roth2020revisiting}.
% Section~\ref{sec:appendix_augmentation} illustrates the details of the multi-view data augmentation strategy.
% Finally, in Section~\ref{sec:appendix_qualitative}, we present more qualitative examples for image retrieval before and after applying the proposed method on the three metric learning benchmarks.

\subsection{Relaxed MS Loss}
\label{sec:relaxed_MS}
The proposed label relaxation technique can be applied to other metric learning losses based on pairwise relations of data. In this section, we present \emph{relaxed MS loss} that is a combination of the label relaxation and Multi-Similarity (MS) loss~\cite{wang2019multi}, the state-of-the-art loss for pair-based metric learning. 
Specifically, relaxed MS loss is obtained by using relaxed relation labels instead of the binary class equivalence indicator and replacing cosine similarity with the relative Euclidean distance, like relaxed contrastive loss in the main paper.
The relaxed MS loss is then formulated as
\begin{align}
\begin{split}
\mathcal{L}(X) =& \frac{1}{n}\sum_{i=1}^{n}\Bigg\{\frac{1}{\alpha}\log \bigg[1 + \sum_{j\neq i}w_{ij}^{s}\exp\Big({\alpha\frac{d_{ij}^t}{\mu_i}\Big)}\bigg] \\
+& \frac{1}{\beta}\log \bigg[1 + \sum_{j\neq i}(1-w_{ij}^{s})\exp\Big({\beta(\delta - \frac{d_{ij}^t}{\mu_i})\Big)}\bigg] \Bigg\},
% &\quad\quad\quad\textrm{where } 
% \mu_i = \frac{1}{n} \sum_{k=1}^{n}{d_{ij}^t}. 
\label{eq:smooth_ms_mu}
\end{split}
\end{align}
where $n$ is the number of samples in the batch, $\delta$ is a margin, and $\alpha>0$ and $\beta > 0$ are scaling factors. Also, $\mu_i = \frac{1}{n} \sum_{k=1}^{n}{d_{ij}^t}$ is the average distance of all pairs associated with $f_i^t$ in the batch.

Table~\ref{tab:emb_transfer_comp} compares relaxed constrastive loss and relaxed MS loss on the three benchmarks for deep metric learning in the \emph{self-transfer} and \emph{dimensionality reduction} setting.
% In Table~\ref{tab:emb_transfer_comp}, this loss is compared to the relaxed contrastive loss on three metric learning benchmarks in the self-transfer and dimensionality reduction setting. 
The details for training are the same as those for relaxed contrastive loss. 
We set $\alpha$ and $\beta$ to 1 and 4 respectively in the self-transfer setting and 1 and 2 respectively in the dimension reduction setting. 
% Interestingly, relaxed MS loss achieves performance comparable to a relaxed contrastive loss in both settings.
As shown in the table, relaxed MS loss achieves performance comparable to a relaxed contrastive loss in both settings.
This result demonstrates the universality of our label relaxation method.
% , can effectively improve performance by applying it to various metric learning loss. 

However, performance of relaxed MS loss is worse than that of the relaxed contrastive loss in most cases, and it demands careful tuning of hyper-parameters for each setting.
In contrast, relaxed contrastive loss is overall better in terms of performance, more robust against hyper-parameter setting, and more interpretable due to its simplicity; this is the reason why we choose relaxed contrastive loss as the representative loss of our framework.

\begin{table*}[!t]
\centering
\begin{tabularx}{\textwidth}{ 
   >{\centering\arraybackslash}X |
   >{\centering\arraybackslash}X |
   >{\centering\arraybackslash}X |
   >{\centering\arraybackslash}X 
   >{\centering\arraybackslash}X
   >{\centering\arraybackslash}X |
   >{\centering\arraybackslash}X
   >{\centering\arraybackslash}X 
   >{\centering\arraybackslash}X |
   >{\centering\arraybackslash}X
   >{\centering\arraybackslash}X
   >{\centering\arraybackslash}X  }
\hline
\multicolumn{3}{l|}{\multirow{2}{*}[-2mm]{Recall@$K$}}         & \multicolumn{3}{c|}{CUB-200-2011} & \multicolumn{3}{c|}{Cars-196} & \multicolumn{3}{c}{SOP}  \\ 
\cline{4-12}
\multicolumn{3}{l|}{}  & 1 & 2 & 4  & 1 & 2 & 4 & 1 & 10 & 100             \\ 
\hline
\multirow{3}{*}{(a)}& \multicolumn{1}{l|}{\emph{Source}: PA~\cite{kim2020proxy}} & \multicolumn{1}{l|}{BN$^{512}$} &69.1 &78.9 &86.1 &86.4 &91.9 &95.0 &79.2 &90.7 &96.2 \\
& \multicolumn{1}{l|}{\ccol Relaxed contrastive} & \multicolumn{1}{l|}{\ccol BN$^{512}$} &\ccol \underline{72.1} &\ccol \underline{81.3} &\ccol \underline{87.6} &\ccol \textbf{89.6} &\ccol\textbf{94.0} &\ccol \textbf{96.5}  &\ccol \textbf{79.8} &\ccol \textbf{91.1} &\ccol\textbf{96.3} \\ 
& \multicolumn{1}{l|}{\ccol Relaxed MS} & \multicolumn{1}{l|}{\ccol BN$^{512}$} &\ccol \textbf{72.3} &\ccol \textbf{81.3} &\ccol \textbf{88.3} &\ccol \underline{89.2} &\ccol\underline{93.9} &\ccol \underline{96.4}  &\ccol \underline{79.3} &\ccol \underline{90.8} &\ccol\underline{96.1} \\ 
\hline

\multirow{3}{*}{(b)} & \multicolumn{1}{l|}{\emph{Source}: PA~\cite{kim2020proxy}} &\multicolumn{1}{l|}{BN$^{512}$} &69.1 &78.9 &86.1 &86.4 &91.9 &95.0 &79.2 &90.7 &96.2 \\ 
& \multicolumn{1}{l|}{\ccol Relaxed contrastive} & \multicolumn{1}{l|}{\ccol BN$^{64}$} &\ccol \underline{67.4}&\ccol \textbf{78.0}&\ccol \textbf{85.9}&\ccol \textbf{86.5}&\ccol \textbf{92.3}&\ccol \textbf{95.3}&\ccol \textbf{76.3}&\ccol \textbf{88.6}&\ccol \textbf{94.8} \\ 
& \multicolumn{1}{l|}{\ccol Relaxed MS} & \multicolumn{1}{l|}{\ccol BN$^{64}$} &\ccol \textbf{67.5}&\ccol \underline{77.9}&\ccol \underline{85.9}&\ccol \underline{86.0}&\ccol \underline{91.5}&\ccol \underline{94.8}&\ccol \underline{75.4}&\ccol \underline{87.9}&\ccol \underline{94.6} \\ 
\hline
\end{tabularx}
\vspace{0.1mm}
\caption{Image retrieval performance of two types of relaxed losses in the two different settings: (a) Self-transfer and (b) dimensionality reduction.  
Embedding networks of the methods are denoted by abbreviations: BN--Inception with BatchNorm.
Superscripts indicate embedding dimensions of the networks.}
\label{tab:emb_transfer_comp}
\vspace{-1mm}
\end{table*}

% ======================================================================
\subsection{Generalization Effect of Label Relaxation}
\label{sec:generalization}

Spectral decay $\rho$~\cite{roth2020revisiting} is a recently proposed generalization measure for deep metric learning.  
It measures KL-divergence between the singular value spectrum of training data embeddings and a uniform distribution. 
Lower $\rho$ value means that a larger number of directions with significant variance exists in the embedding space, thus indicates better generalization~\cite{roth2020revisiting}.

\begin{figure} [!t]
\centering
\includegraphics[width = 0.97\columnwidth]{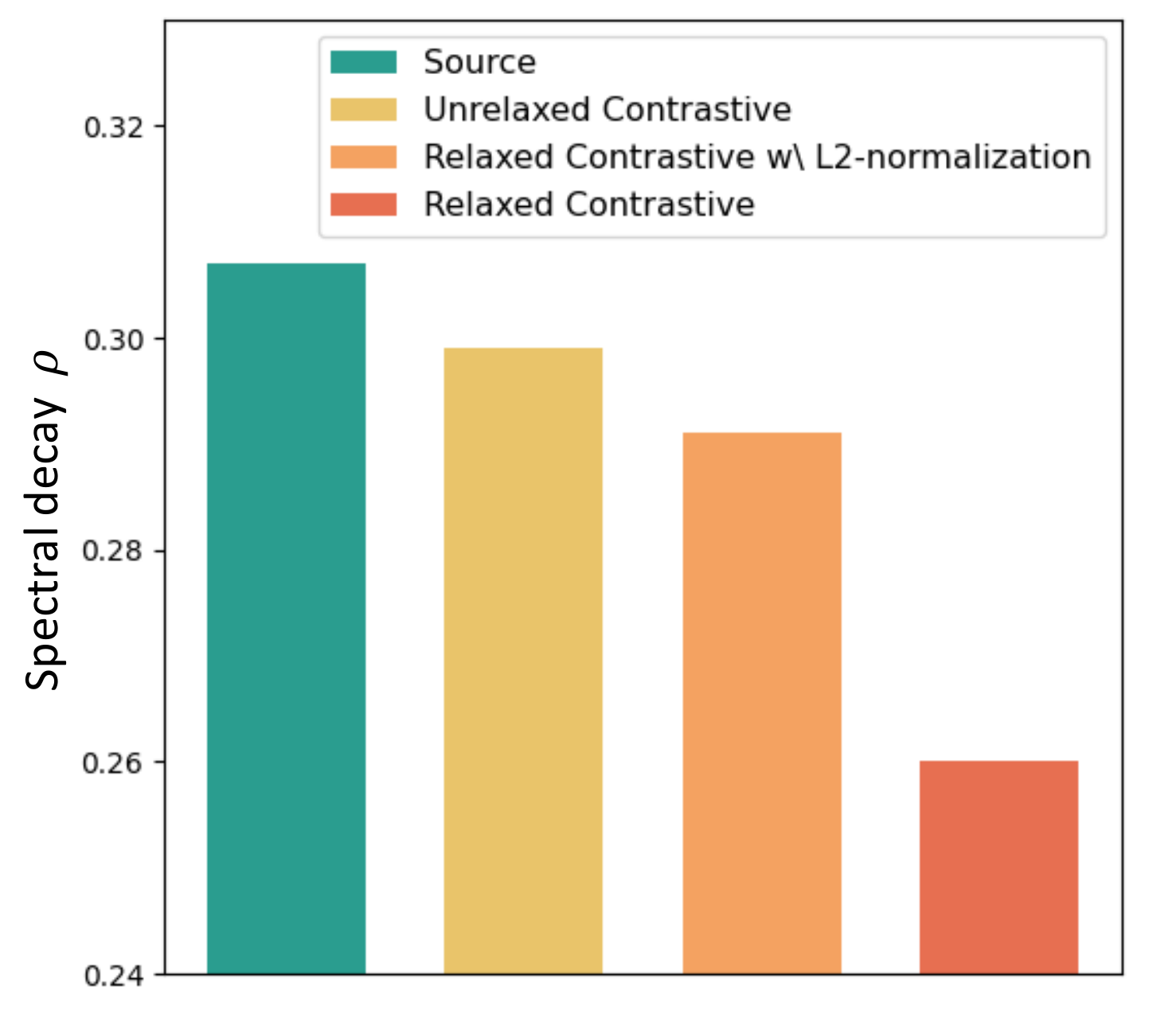}
\caption{Spectral decay $\rho$ of source and target embedding models trained on the Cars-196 dataset.} 
\label{fig:sup_rho}
\end{figure}

In Fig.~\ref{fig:sup_rho}, Spectral decay $\rho$ of the source and target embedding models is presented. Target embedding models are trained with our method and its variants.
As shown in the figure, target embedding models after embedding transfer have lower $\rho$ value than the source, and our method significantly reduces $\rho$ value compared to its unrelaxed or L2-normalized version.
We argue that our method reduces $\rho$ value and improves generalization performance since the relaxed relation labels and the relative pairwise distance helps target embedding space to encode rich pairwise relation without restriction on the manifold.
% , our method reduces $\rho$ value and improves generalization performance.
% ======================================================================

\begin{table}[!t]
\centering
\begin{tabularx}{0.95\columnwidth}{ 
   >{\centering\arraybackslash}X |
   >{\centering\arraybackslash}X |
   >{\centering\arraybackslash}X} 
\hline
\multicolumn{1}{l|}{\multirow{2}{*}[0mm]{Methods}} & \multicolumn{2}{c}{Recall@1} \\  \cline{2-3}
& CUB& Cars\\ \hline
 \multicolumn{1}{l|}{\ccol Ours} & \ccol 72.1 & \ccol 89.6\\ 
 \multicolumn{1}{l|}{Ours w/o augmentation} &  71.5 & 89.0\\ \hline
\end{tabularx}
\vspace{2mm}
\caption{
% Comparison of source embedding and ablations of our method on the CUB-200-2011 (CUB) and Cars-196 (Cars) datasets.
Effect of multi-view augmentation strategy on the CUB-200-2011 (CUB) and Cars-196 (Cars) datasets. 
}
\label{tab:augmentation}
% \vspace{-4mm}
\end{table}

\subsection{Details of Multi-view Data Augmentation}
\label{sec:appendix_augmentation}

\begin{figure*} [!h]
\centering
\includegraphics[width = 0.9\textwidth]{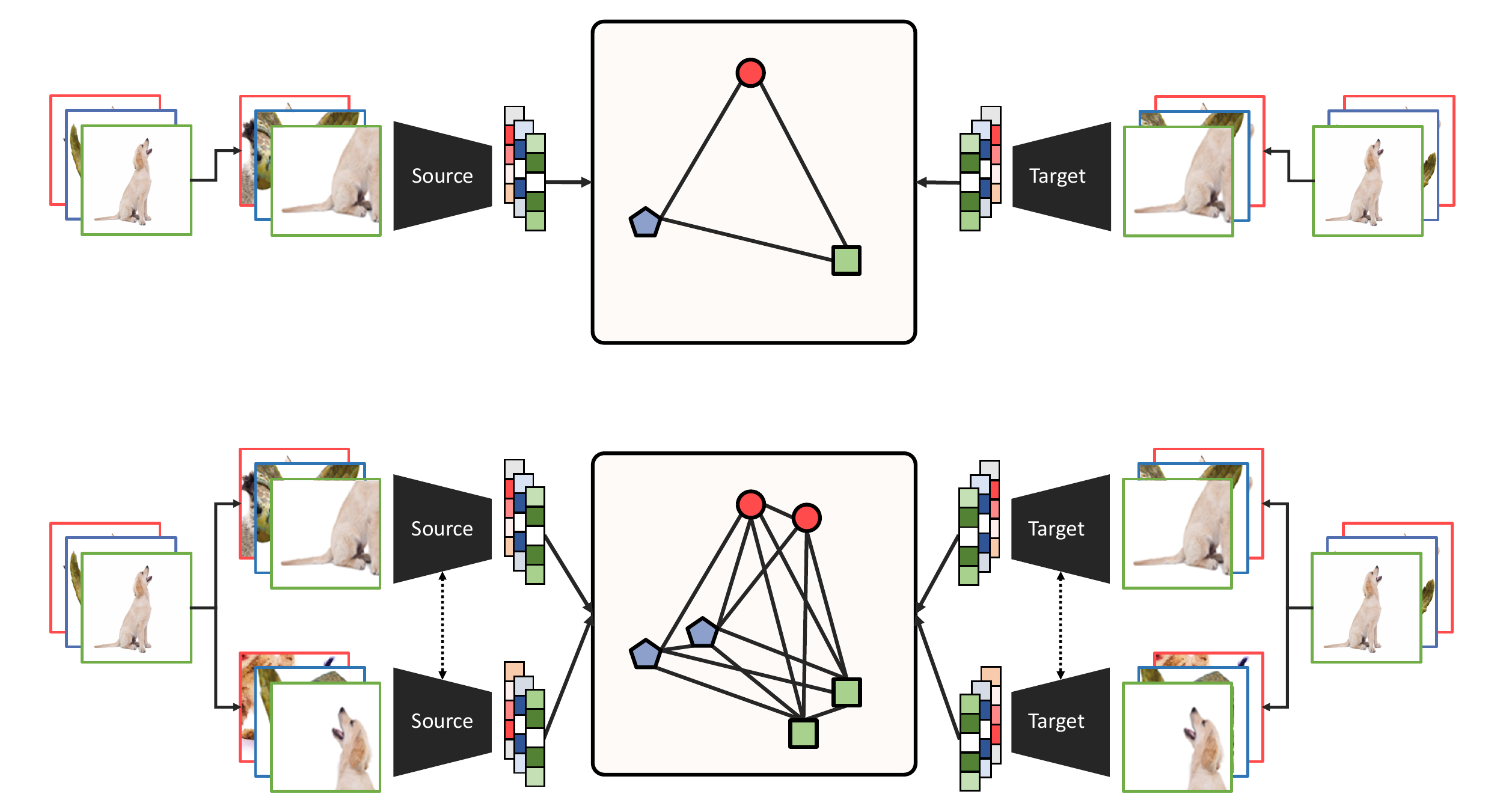}
\caption{Comparison of standard data augmentation (\emph{top}) and our multi-view augmentation (\emph{bottom}). Different colors and shapes represent distinct samples.} 
\label{fig:sup_multiview}
\end{figure*}

In recent approaches to self-supervised representation learning, the use of multi-view samples produced from the same image plays an important role for performance improvement. 
We present a simple yet effective multi-view augmentation strategy for enhancing the effect of embedding transfer. It helps transfer knowledge by considering relations between multiple views of individual samples, such as relations between different parts of an object. 

The overall procedure of our multi-view augmentation is as follows. 
We first apply the standard random augmentation technique multiple times to images of input batch.
Then, all augmented multi-view images are passed through the source and target embedding networks. 
Note that the source and target model take the same augmented image as input. 
The output embedding vectors are concatenated and used as the inputs of the embedding transfer loss. Fig.~\ref{fig:sup_multiview} illustrates this procedure where the number of views is two. 
The top and bottom of the figure describe the standard augmentation technique and our strategy, respectively. When using standard augmentation, only relations between different samples are considered. 

Applying a multi-view augmentation strategy for embedding transfer allows knowledge transfer to consider more diverse and detailed relations between samples produced from the same image.
The empirical advantage of the multi-view augmentation is verified in Table~\ref{tab:augmentation}, where it improves the stability and convergence of embedding transfer as well as the performance of target embedding models. 

% ======================================================================
\subsection{Impact of Hyperparameters}
\label{sec:hyperparameters}

\begin{figure} [!t]
\centering
\includegraphics[width = 0.9\columnwidth]{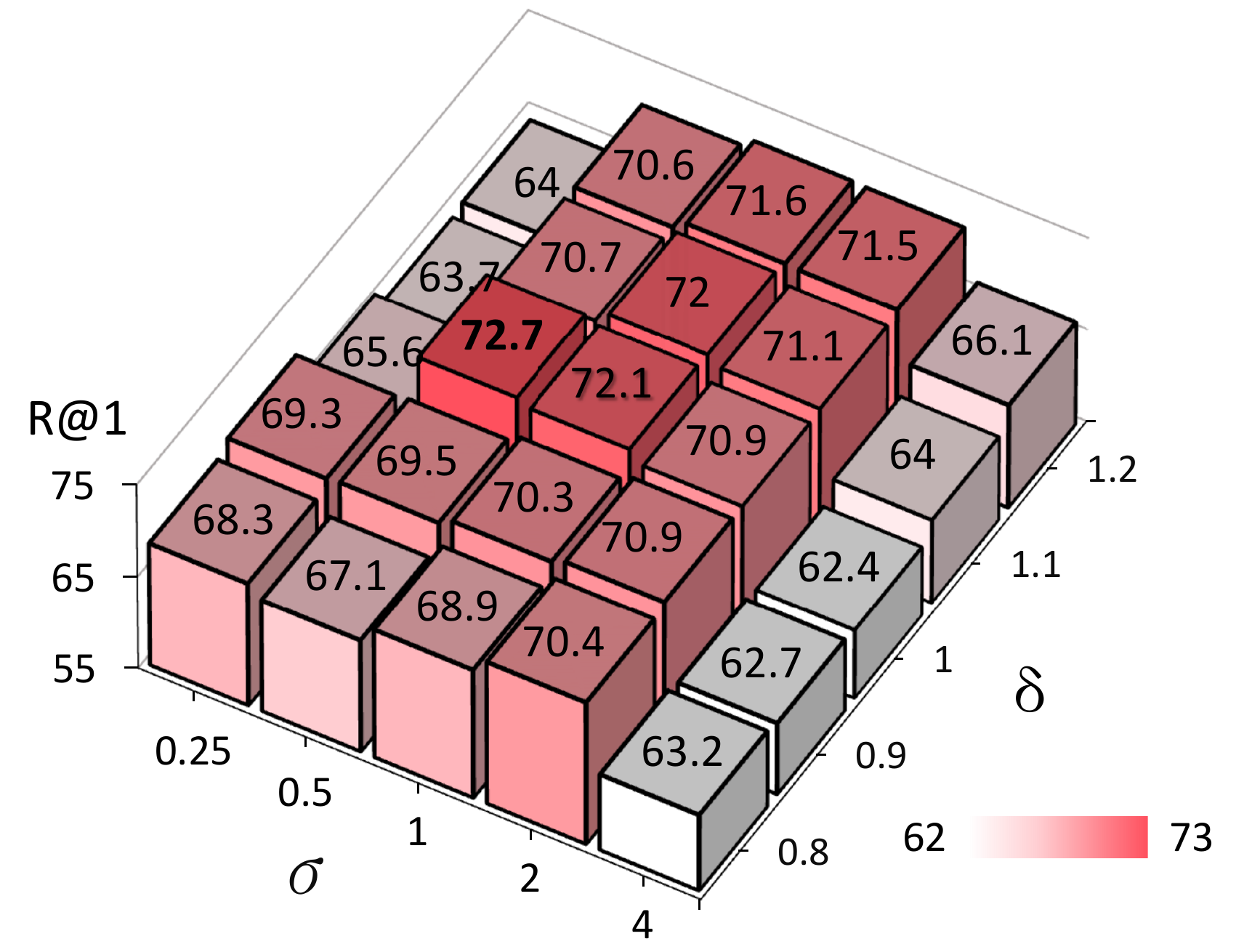}
\caption{Recall@1 versus hyperparameters $\delta$ and $\sigma$ on the CUB dataset.} 
\label{fig:sensitivity}
\end{figure}

We empirically investigated the effect of the hyperparameters $\delta$ and $\sigma$ on performance. We examine Recall@1 in accuracy of relaxed contrastive loss by varying the values of the hyperparameters $\sigma \in \{0.25, 0.5, 1, 2, 4\}$ and $\delta \in \{0.8, 0.9, 1, 1.1, 1.2\}$.
As summarized in Figure~\ref{fig:sensitivity}, the accuracy of our method was consistently high and outperformed state of the art in most cases when $\sigma$ is greater than 0.25 and less than 4.
Note that the values of $\delta$ and $\sigma$ used in the paper are not optimal as we did not tune them using the test set.
% ======================================================================

% ======================================================================
\subsection{Additional Qualitative Results} % for Deep Metric Learning}
\begin{figure*} [!t]
\centering
\includegraphics[width = \textwidth]{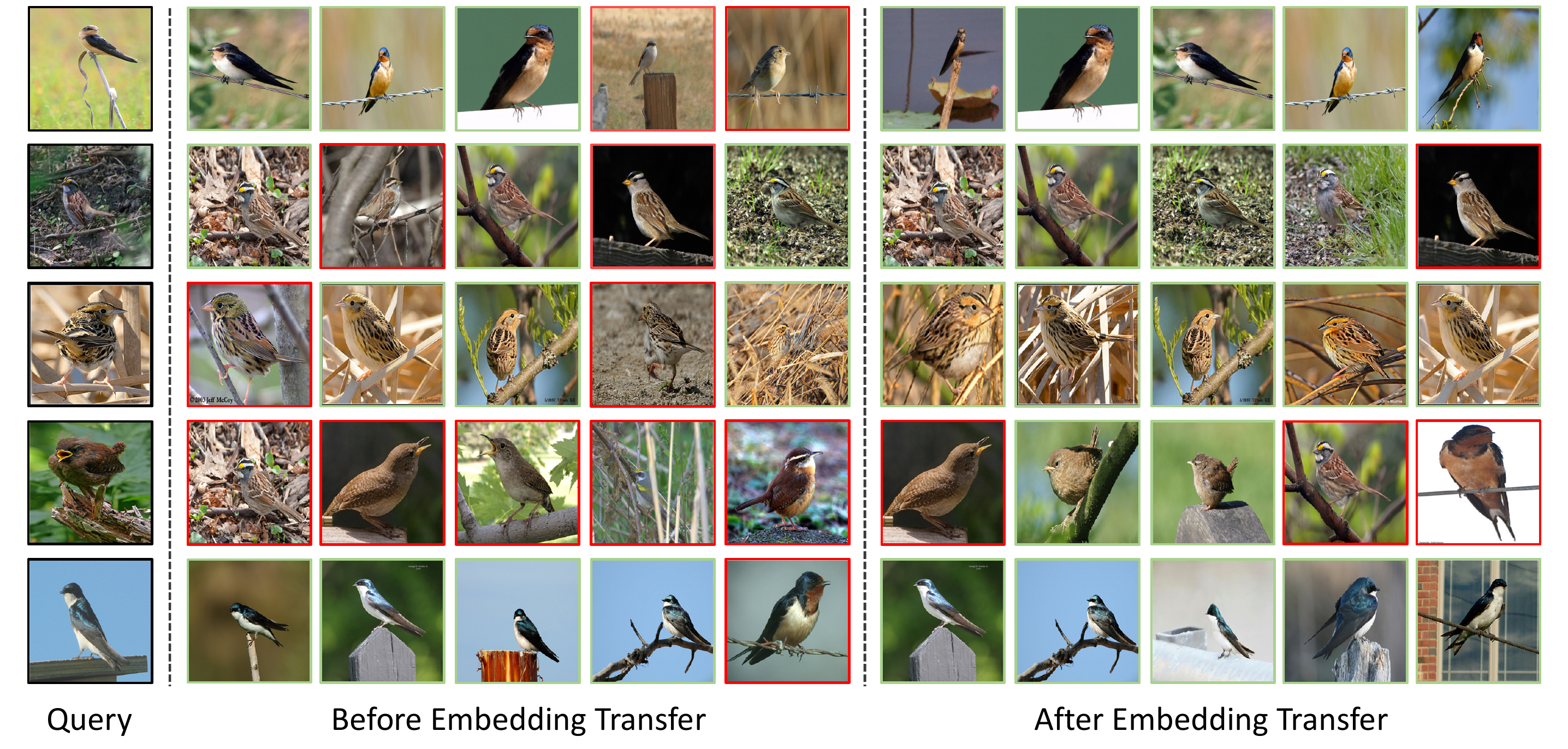}
\caption{Top 5 image retrievals of the state of the art~\citep{kim2020proxy} before and after the proposed method is applied on the CUB-200-2011 dataset. Images with green boundary are success cases and those with red boundary are false positives. 
} 
\label{fig:sup_qualitative_results_cub}
\end{figure*}

\begin{figure*} [!t]
\centering
\includegraphics[width = \textwidth]{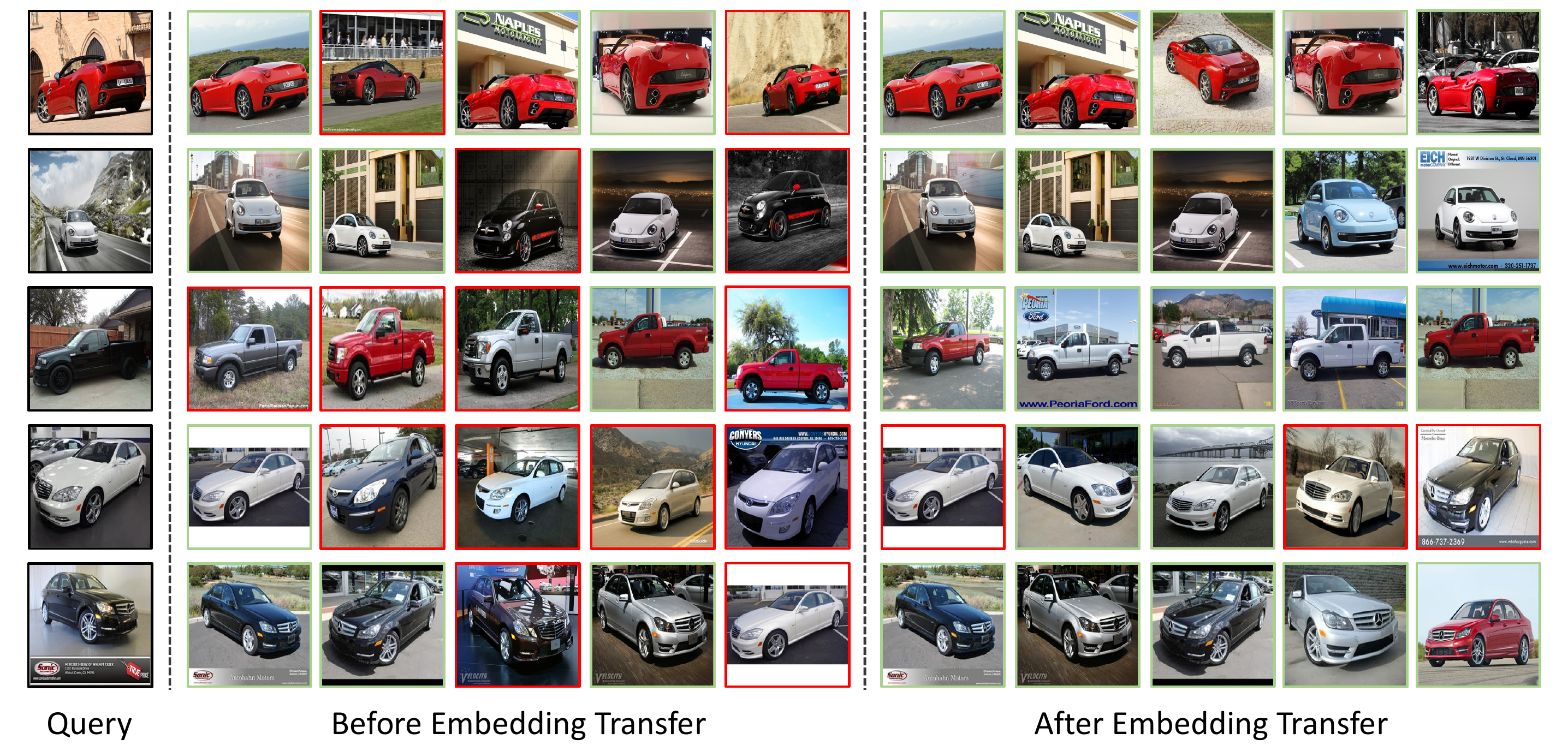}
\caption{Top 5 image retrievals of the state of the art~\citep{kim2020proxy} before and after the proposed method is applied on the Cars-196 dataset. Images with green boundary are success cases and those with red boundary are false positives.
} 
\label{fig:sup_qualitative_results_cars}
\end{figure*}

\begin{figure*} [!t]
\centering
\includegraphics[width = \textwidth]{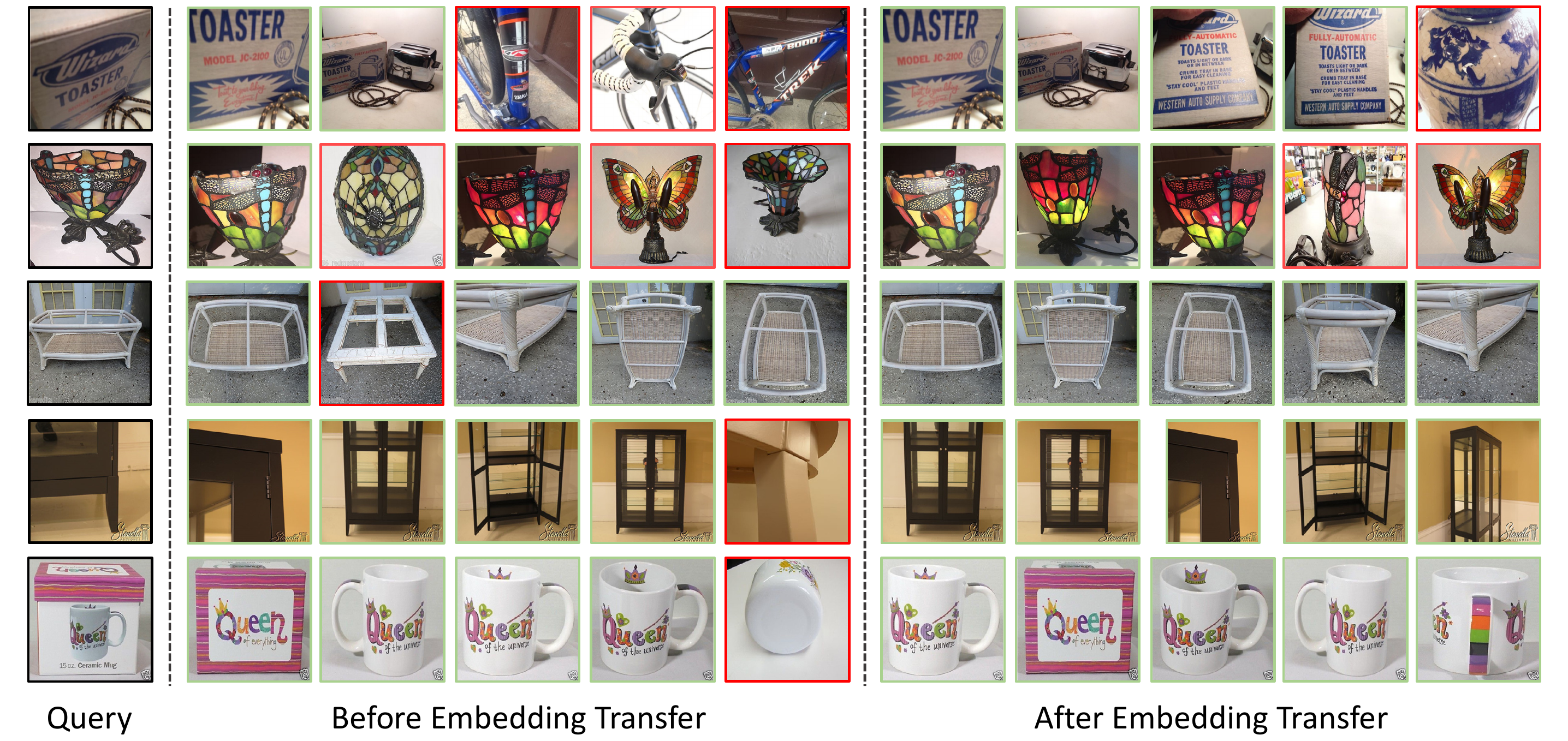}
\caption{Top 5 image retrievals of the state of the art~\citep{kim2020proxy} before and after the proposed method is applied on the SOP dataset. Images with green boundary are success cases and those with red boundary are false positives.
} 
\label{fig:sup_qualitative_results_sop}
\end{figure*}

\label{sec:appendix_qualitative}
More qualitative results of image retrieval on the CUB-200-2011, Cars-196, and SOP datasets are presented in Fig.~\ref{fig:sup_qualitative_results_cub}, \ref{fig:sup_qualitative_results_cars}, and~\ref{fig:sup_qualitative_results_sop}, respectively. 
We prove the positive effect of the proposed method by showing qualitative results before and after applying the proposed method in the \emph{self-transfer} setting;
the source embedding model is Inception-BatchNorm with 512 embedding dimension and trained with the proxy-anchor loss~\citep{kim2020proxy}.
% to the source model trained with proxy-anchor loss~\citep{kim2020proxy}. 
The overall results indicate that the proposed method significantly improves the source embedding model. 
From the examples of the 2nd, 3rd, and 5th rows of Fig.~\ref{fig:sup_qualitative_results_cub},
both models retrieve birds visually similar to the query, but only the models after embedding transfer successfully retrieved birds of the same species. 
Meanwhile, the examples of the 2nd, 3rd, and 5th rows of Fig.~\ref{fig:sup_qualitative_results_cars} show that the model trained with our method provides accurate results regardless of the color changes of the cars.
Also, in the examples of the 2nd and 3rd rows of Fig.~\ref{fig:sup_qualitative_results_sop},
the source model makes mistakes easily since the false positives are similar to the query in terms of appearance, yet it becomes more accurate after applying to embedding transfer with the proposed method.

\end{document}